\pdfoutput=1
\documentclass[11pt]{article}
\usepackage[]{acl}
\usepackage{times}
\usepackage{latexsym}
\usepackage{times}
\usepackage{latexsym}
\usepackage[T1]{fontenc}
\usepackage[utf8]{inputenc}
\usepackage{inconsolata}
\usepackage{graphicx}
\usepackage{setspace}
\usepackage{subcaption}
\usepackage{times}
\usepackage{setspace}
\usepackage{amsmath,amsthm,amsfonts,amssymb,bm}
\usepackage{latexsym}
\usepackage{hyperref}
\usepackage{inconsolata}
\usepackage{url}
\usepackage{listings}
\usepackage{multirow}
\usepackage{arydshln} 
\usepackage{enumitem}
\usepackage{caption}
\usepackage{colortbl}
\usepackage[most]{tcolorbox}
\usepackage{subcaption}
\usepackage{titletoc}
\usepackage[capitalize]{cleveref} 
\usepackage{tabularx}
\usepackage{booktabs}
\usepackage{multirow}
\usepackage{tabularx}
\usepackage{array} 
\usepackage{amsmath}
\usepackage{adjustbox}
\usepackage{booktabs} 
\usepackage{bm}
\usepackage[normalem]{ulem}
\useunder{\uline}{\ul}{}
\definecolor{lightblue}{HTML}{DAE8FC}
\definecolor{green}{HTML}{D5E8D4}
\definecolor{orange}{HTML}{FFE6CC}
\usepackage{afterpage}
\usepackage{microtype}
\usepackage{footmisc}

\title{From Generic Empathy to Personalized Emotional Support: A Self-Evolution Framework for User Preference Alignment}

\author{
    Jing Ye$^{1,2}$, 
    Lu Xiang$^{1,2}$\Thanks{ Corresponding Author},
    Yaping Zhang$^{1,2}$,
    Chengqing Zong$^{1,2}$\\
    \footnotesize${}^1$State Key Laboratory of Multimodal Artificial Intelligence Systems, Institute of Automation, CAS, Beijing, China\\
    \footnotesize${}^2$School of Artificial Intelligence, University of Chinese Academy of Sciences, Beijing, China\\ 
    \footnotesize{yejing2022@ia.ac.cn}; \footnotesize{\{lu.xiang, yaping.zhang,cqzong\}@nlpr.ia.ac.cn} \\
}

\begin{document}
\maketitle
\begin{abstract}

Effective emotional support hinges on understanding users' emotions and needs to provide meaningful comfort during multi-turn interactions. 
Large Language Models (LLMs) show great potential for expressing empathy; however, they often deliver generic and one-size-fits-all responses that fail to address users' specific needs. To tackle this issue, we propose a self-evolution framework designed to help LLMs improve their responses to better align with users' implicit preferences concerning user profiles (personalities), emotional states, and specific situations.
Our framework consists of two distinct phases: \textit{(1)} \textit{Emotional Support Experience Acquisition}, where LLMs are fine-tuned on limited emotional support conversation data to provide basic support, and \textit{(2)} \textit{Self-Improvement for Personalized Emotional Support}, where LLMs leverage self-reflection and self-refinement to generate personalized responses. Through iterative direct preference optimization between the pre- and post-refined responses, our model generates responses that reflect a better understanding of the user's implicit preferences. Extensive experiments and evaluations demonstrate that our method significantly enhances the model's performance in emotional support, reducing unhelpful responses and minimizing discrepancies between user preferences and model outputs.  

\end{abstract}

\section{Introduction}

\begin{figure}[t]
    \centering
    \includegraphics[width=\linewidth]{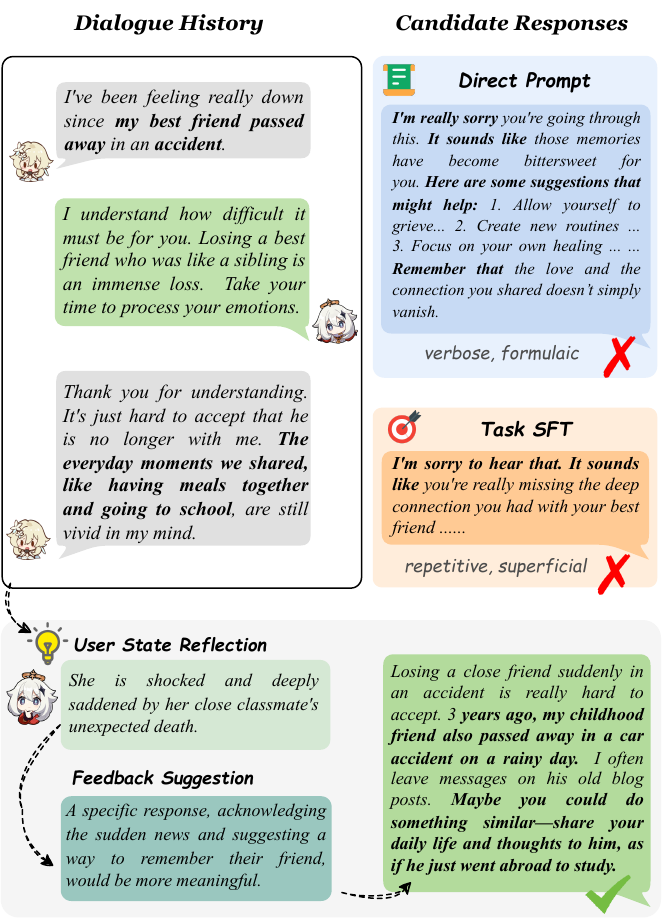}
    \caption{Example responses. \textit{Direct prompting of LLaMA} results in verbose and formulaic outputs. \textit{Task-Specific SFT} is empathetic but often lacks depth and variety, giving it a perceived "AI-like" quality. In contrast, \textit{self-reflection} on user preferences provides a pathway to more specific and engaging responses.}
    \vspace{-5mm}
    \label{fig:demo_intro}
    \vspace{-1mm}
\end{figure}

Emotional support conversation (ESC) systems require a deep understanding of users' emotions and need to provide meaningful comfort and assistance during multi-turn interactions \cite{ijcai/00080XXSL22, rains2020support}, which are vital in practical applications such as mental health care, emotional companionship, and customer service. Given that each user has unique emotional needs and experiences \cite{rogers2013client}, delivering personalized and contextually appropriate emotional support is essential for ensuring practical assistance \cite{campos2018challenges, pal}.

Despite the promising potential of LLMs for generating empathetic responses \cite{llama2, qwen2, gpt-4}, they often struggle to provide diverse and contextually appropriate support \cite{Muffin}. As illustrated in Figure \ref{fig:demo_intro}, \textbf{direct prompting LLMs often results in superficial empathy, verbosity, and formulaic structures}. 
A simple yet effective approach is supervised fine-tuning (SFT) LLMs on ESC corpora \cite{ExTES, AugESC, SMILE, ESCoT}. However, SFT relies on substantial, high-quality ESC data, which is often scarce and difficult to acquire. Moreover, \textbf{over-reliance on SFT can lead to repetitive responses that express empathy overtly but lack depth and variety }\cite{abs-2303-06135}. As demonstrated in Figure \ref{fig:demo_intro} and \ref{fig:combined_wordclouds}, SFT models can fall into predictable patterns, frequently using phrases like "\textit{It sounds like…}" or "\textit{I'm sorry to hear that…}".

Recent insights highlight that LLMs can self-improve their performance through self-reflection and self-refinement guided by human-designed principles \cite{lu2024selfselfevolutionlanguagefeedback, Self-Refine, selfee2023, yasunaga2024almaalignmentminimalannotation}. Inspired by these findings, we pose the intriguing question: \textit{{\ul Can LLMs be taught to consider what kind of responses are genuinely needed by users, and can this reflective process lead to refined and more personalized responses?}} 

This work seeks to bridge the gap between generic empathetic responses and truly user-centered personalized emotional support by incorporating self-reflection and self-refinement into automated systems. Effective ES systems require an iterative approach that continuously reflects ongoing dialogue to refresh user understanding and refine responses, ultimately delivering targeted empathy and tailored solutions.
The empirical evidence presented in Figure \ref{fig:demo_intro} and Table \ref{tab:ablation study on llama} demonstrates that instructing LLMs to summarize user situations, infer emotions and causes, and choose appropriate support strategies leads to a significant improvement in response quality.

\begin{figure*}
    \centering
    \includegraphics[width=1\linewidth]{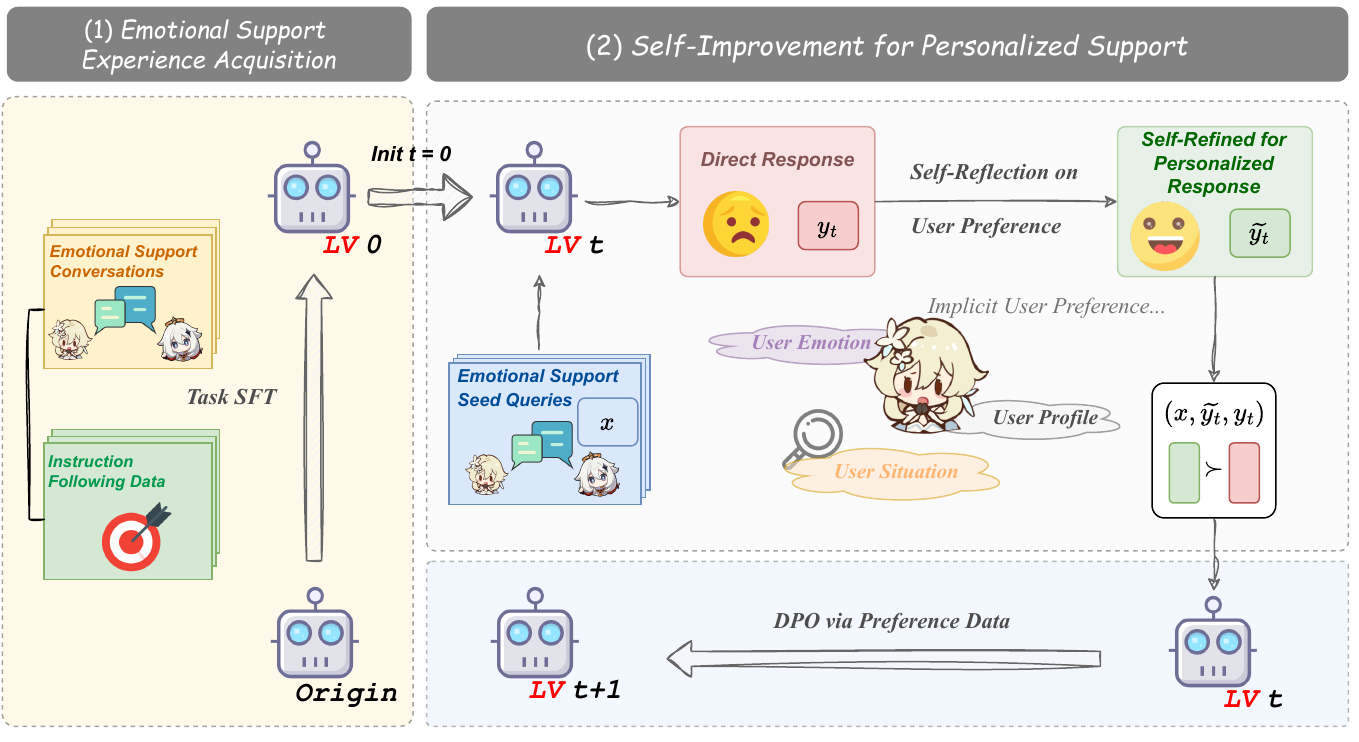}
    \vspace{-5mm}
    \caption{The overview of our self-evolution framework, which enhances personalized emotional support capabilities through a two-stage learning phase: (1) Emotional Support Experience Acquisition: We fine-tune LLMs on minimal human-annotated ESC data, equipping them with basic emotional support capability. (2) Self-Improvement for Personalized Emotional Support: We utilize the LLMs' self-reflection abilities to tailor responses to the user's personality, situation, and emotions. The pre- and post-refined responses are natural synthetic preference data. The process involves iterative preference optimization for generating responses that align with the user's implicit preferences, eliminating the need for explicit reflection steps.
}
    \label{fig:overview}
    \vspace{-2mm}
\end{figure*}

To this end, we introduce a self-evolution framework for user preference alignment. As depicted in Figure \ref{fig:overview}, our self-evolution framework comprises two steps: (1) \textit{Emotional Support Experience Acquisition}: we first fine-tune LLMs on limited ESC data, enabling them to provide essential emotional support. (2) \textit{Self-Improvement for Personalized Emotional Support}: Subsequently, we leverage LLMs' inherent self-reflection and self-refinement capabilities to generate responses that consider the implicit user preference, including \textit{profile}, \textit{situation}, and \textit{emotions}. The pre- and post-refined responses are considered the preference data. Through direct preference optimization, the model generates responses that reflect an understanding of the user's implicit preferences during interactions, thereby eliminating the need for explicit reflection and refinement steps. Experimental results and extensive human evaluations indicate that our generated responses are more diverse and better aligned with user input. These improved responses effectively reduce ineffective empathy and preference misalignment, facilitating more productive multi-turn interactions.

Our main contributions can be summarized as follows:
\begin{itemize}[itemsep= 0.1pt,topsep = 0.1pt,partopsep=0.1pt]
\item We reveal the limitations of the current Emotional Support Chatbot, which is notably deficient in understanding users' implicit preferences, resulting in repetitive and superficial expressions of empathy.
\item We present a simple yet effective self-evolution framework for personalized emotional support without explicit reflection and refinement steps. 
\item Experimental results and comprehensive human evaluations demonstrate that our method effectively minimizes unhelpful responses and discrepancies in personalized preferences.
\end{itemize}

\section{Method}

Inspired by recent insights highlighting that LLMs can self-improve through language feedback \cite{lu2024selfselfevolutionlanguagefeedback, Self-Refine, selfee2023, yasunaga2024almaalignmentminimalannotation}, we present a self-evolution framework designed to enable LLMs to provide personalized emotional support.  
This framework operates in two phases: Emotional Support Experience Acquisition (Section \ref{pharse1}) and Self-Improvement for Personalized Emotional Support (Section \ref{pharse2}).

\subsection{Emotional Support Experience Acquisition}
\label{pharse1}

\subsubsection{Task Definition}
Emotional Support (ES) involves understanding the user's situation and choosing appropriate supportive strategies to alleviate their distress. Formally, we can denote the ES model as $\mathcal{M}$, and represent the current dialogue context as $\mathcal{C}_n = (q_{1}, r_{1}, ... , q_{i-1}, r_{i-1}, ... q_{n})$. In this representation, $q_i$ and $r_i$ correspond to the $i$-th utterance from the user and the model, respectively. Given the task and strategy description prompt $\mathcal{P}_{task}$, the goal of the ES model is to generate an emotional supportive response $r_n$, which can be represented as:
\begin{equation}
    r_n = \mathcal{M}(\mathcal{P}_{task} \ || \ \mathcal{C}_n)
\end{equation}

\subsubsection{Task Learning}
We equip the model with emotional support capability by fine-tuning the backbone on the manually annotated ESConv dataset \cite{ESConv}. To preserve the general abilities, we employ Low-Rank Adaptation \cite{LoRA}, fine-tuning only the LoRA adapter parameters. We further incorporate a replay mechanism by incorporating some instruction-following data \cite{wang-etal-2024-inscl}. The model is trained using the SFT loss:
\begin{equation} 
    \mathcal{L}_{SFT} = - \text{log} P(\bm{y} | \bm{x}, \mathcal{P}; \theta)
\end{equation}
where $\bm{x}, \bm{y}$ represent the input and output of the model, respectively, while $\mathcal{P}$ denotes the task description or instructions. The resulting fine-tuned model is denoted as $\mathcal{M}^0$.

\subsection{Self-Improvement for Personalized Emotional Support}
\label{pharse2}
While SFT improves empathetic response generation, it often produces superficial outputs, failing to capture nuanced user preferences that are crucial for effective emotional support. To address this limitation and minimize unhelpful responses, we introduce a self-improvement method based on iterative direct preference optimization (DPO) \cite{dpo}. Guided by human design principles, the model reflects on the user's personality, situation, and emotion to refine its responses. These pre- and post-refined responses naturally serve as rejected and chosen candidates, respectively \cite{synpo}. Through direct preference optimization, the model generates responses that reflect an understanding of the user's implicit preferences during interactions, thereby eliminating the need for explicit reflection and refinement steps.

\subsubsection{Synthetic Preference Data Generation}
\noindent \textbf{Rejected Response Generation}
Constructing high-quality preference data pairs requires a diverse set of user queries. While synthetic ESC datasets may not produce emotional support responses comparable to human quality, they offer a valuable source of varied queries \cite{ExTES, AugESC}. We extract the dialogue context $\mathcal{C}_n$ from these synthetic datasets, where $n$ is the turn index, and employ $\mathcal{M}^t$ to generate responses. 
\begin{equation} 
    y^t_n = \mathcal{M}^t(\mathcal{C}_n), \quad \text{initial } t=0
\end{equation}
These unrestricted responses are treated as rejected responses. 

\noindent \textbf{Self-Reflection on Implicit User Preference}
Research indicates that LLMs possess strong contextual inference capabilities \cite{qwen2, llama3}, enabling them to infer user emotions, implicit profiles, and even personality from ongoing conversations. Given the dialogue history $\mathcal{C}_n$ and human-designed principles $\mathcal{I}$, the model $\mathcal{M}^t$ is tasked with summarizing the user's profile $u_n$ and current emotional state $s_n$ according to the following equation:
\begin{equation} 
    (u_n, s_n) = \mathcal{M}^t(\mathcal{I} \ || \ \mathcal{C}_n) 
\end{equation}
$u_n$ and $s_n$ are continuously updated throughout the conversation, enabling the model to refine its understanding of the user.

\noindent \textbf{Self-Refinement for Personalized Responses}
Responses generated solely from dialogue history often fail to capture the user's implicit preferences. Drawing on insights from psychological research, user preferences can be decomposed into two key dimensions: long-term traits, encapsulated by the user profile \cite{fleeson2001toward}, and context-sensitive emotional needs \cite{article}.
To better understand and adapt to these implicit preferences, we leverage the strong contextual reasoning of LLMs \cite{qwen2, llama3}.  Given the dialogue history $\mathcal{C}_n$ and human-designed principles $\mathcal{I}$, the model $\mathcal{M}^t$ is tasked with summarizing the user's profile $p_n$ and current emotional state $s_n$ according to the following equation:
\begin{equation} 
    \tilde{y^t_n} = \mathcal{M}^t(\mathcal{I} \ || \ \mathcal{C}_n, u_n, s_n, y^t_n)
\end{equation}
The pre- and post-refined responses form a preference pair $(y_t,\tilde{y^t_n})$, serving as the rejected and chosen candidates, respectively. Inevitably, some low-quality data is generated during this process; the data filtering process is detailed in Appendix \ref{data filtering}, and the prompts for self-reflection and self-refinement are shown in Figure \ref{prompts for ablation}.

\subsubsection{Preference Optimization}
The synthetic preference data generation process naturally facilitates iterative self-improvement. In each iteration, we employ DPO \cite{dpo} for training. 
\begin{equation}
\begin{split}
    \mathcal{L}_{DPO} &= \log \sigma (\beta \cdot \log \frac{P( \tilde{y^t_n} | \mathcal{C}_n; \theta)}{P(\tilde{y^t_n} | \mathcal{C}_n; \theta')}  \\[5pt]
    &\quad \quad \quad - \beta \cdot \log \frac{P( y^t_n | \mathcal{C}_n; \theta)}{P( y^t_n | \mathcal{C}_n; \theta')})
\end{split}
\end{equation}
To mitigate the instability of DPO training, we incorporate an SFT loss on the chosen responses during optimization.
\begin{equation}
\begin{split}
    \mathcal{L}_{SFT} &= -\log P( \tilde{y^t_n} | \mathcal{C}_n; \theta)
\end{split}
\end{equation}
The final optimization loss is:
\begin{equation}
\begin{split}
    \mathcal{L} &= \mathcal{L}_{DPO} + \gamma \cdot \mathcal{L}_{SFT} 
\end{split}
\end{equation}
here $\beta$ and $\gamma$ are set to $0.1$ and $1$, respectively.

\section{Experiments}
\subsection{Dataset}

We collect three ESC datasets: the manually annotated ESConv dataset \cite{ESConv}, and the synthetically generated ExTES \cite{ExTES} and ServeForEmo \cite{ServeForEmo}. Detailed statistics are available in Table \ref{dataset}. ESConv is split into training and testing sets with a 9:1 ratio. During the \textit{Emotional Support Experience Acquisition}  stage, we use the ESConv training set along with 500 instruction-following samples from Alpaca \cite{alpaca}. And we combine ExTES and ServeForEmo as seed data for synthetic preference data generation.

\begin{table}[t]
\centering
\small
\setstretch{1.2}
\begin{adjustbox}{width=\columnwidth}
\begin{tabular}{lccc}
\toprule[1.2pt]
\multicolumn{1}{c}{\textbf{Dataset}} & \textbf{ExTES} & \textbf{ESConv} & \textbf{ServeForEmo} \\ \midrule
\# Session                           & 11,167         & 1,295           & 3,749                          \\
Avg Session Len                   & 16.68          & 22.58           & 15.91                      \\
Avg Utter. Len                    & 29.59          & 21.17           & 18.45                         \\
Avg Seeker Utter. Len                  & 22.63          & 19.90           & 15.39                        \\
Avg Supporter Utter. Len                & 36.55          & 22.44           & 21.51                        \\ 
\bottomrule[1.2pt]
\end{tabular}
\end{adjustbox}
\caption{The Statistics of Emotional Support Datasets. Conversations in these datasets typically span seven turns, with an average utterance length of approximately 20 words.}
\label{dataset}
\end{table}

\subsection{Implementation Details}
This study employs three frequently used LLMs as backbones: LLaMA-3-8B-Instruct\footnote{https://huggingface.co/meta-llama/Meta-Llama-3-8B-Instruct}, Qwen2-7B-Instruct\footnote{https://huggingface.co/Qwen/Qwen2-7B-Instruct}, and Mistral-7B-Instruct-v0.3\footnote{https://huggingface.co/mistralai/Mistral-7B-Instruct-v0.3}. The LoRA technique \cite{LoRA} is employed across all experiments, featuring a LoRA adapter with a rank of 8 and alpha of 16 into each linear module. 
For optimization, we utilize the AdamW optimizer \cite{adamw} with a learning rate of $5 \times 10^{-6}$ and a linear warm-up during the initial 1\% of the training steps. The batch size is set to 4 per device, with gradient accumulation every two steps across two epochs. Early stopping is implemented with a patience threshold of 3 evaluation steps to mitigate over-fitting. For generation and evaluation, we set the decoding parameters to a temperature of 0.9, top-p of 0.8, top-k of 50, and a repetition penalty of 1.2.
All experiments are conducted on 1 NVIDIA L40 40GB GPU. The implementation framework utilized is \text{LLaMA-Factory} \cite{zheng2024llamafactory}.

\subsection{Baselines}
To evaluate the effectiveness of our approach, we conducted a comparative evaluation across three categories under identical experimental settings:

\noindent \textbf{Vanilla}: Instruction-based backbone models provided with ESC task prompts. These served as baselines to assess inherent capabilities without task-specific fine-tuning.

\noindent \textbf{SFT}: LLMs fine-tuned on two dataset types: the ESConv dataset (\textbf{\textit{SFT-ESConv}}) and synthetic ESC datasets including ExTES and ServeForEmo (\textbf{\textit{SFT-SynESC}}).

\noindent \textbf{Self-Evolution with Preference Learning}: Models at different iterations in our self-evolution framework: 
\begin{itemize}[itemsep= 0pt,topsep = 0.1pt,partopsep=0.1pt]
    \item $\mathcal{M}^0$: The initial fine-tuned ES model.
    \item $\mathcal{M}^t$: Models initialized from $\mathcal{M}^{t-1}$ and optimized using synthetic preference data generated by $\mathcal{M}^{t-1}$.
\end{itemize}

\subsection{Evaluation Details}
\subsubsection{Evaluation Settings}
\label{evaluation settings}
Our evaluation comprises \textbf{objective} and \textbf{subjective} assessments. 
The objective evaluation measures the similarity between model-generated and manually annotated responses using the ESConv test set. 
Recognizing the limitations of text overlap metrics for the open-ended ES task, which can penalize informative and creative responses, we prioritize subjective evaluation to better reflect real-world user experience. This subjective assessment incorporates interactive pointwise and pairwise human evaluations. 
Appenix \ref{human evaluation interface} illustrate the  evaluation process and guidelines, respectively.

\noindent \textbf{Interactive Pointwise Evaluation}: 

To mitigate evaluation bias, we employ an interactive pointwise evaluation where dialogue sessions were randomly assigned to different models. Participants, consisting of 50 undergraduate students with diverse backgrounds, rate their satisfaction with the assigned ES agent on a 5-point Likert scale \cite{Likert} across predefined dimensions. Higher scores indicate better performance. The final score for each model is calculated by averaging the ratings across all participants. Each dialogue includes at least eight turns. LLM-as-a-judge pointwise evaluations are also provided in the Appendix \ref{llm-as-judge-results}.

\noindent \textbf{Interactive Pairwise Evaluation}: 
Four graduate students engage in dialogues with the models, with each dialogue lasting at least ten turns. At each turn, two models (A and B) generate responses simultaneously based on user input. The user then selects "A win", "B win", or "tie". The winning response is appended to the dialogue history for subsequent turns \cite{zhou-etal-2024-characterglm}. In the event of a tie, the user can choose to continue the conversation with either response.

\subsubsection{Evaluation Metrics}
\label{evaluation metrics}

\paragraph{Automation Evaluation} We employ five established automatic evaluation metrics. BLEU-n\cite{bleu}, ROUGE-L \cite{rouge}, METEOR\cite{meteor}, and BERT-Score \cite{BERTScore} metrics are used to assess similarity with the human-written references. For evaluating diversity, Distinct-n \cite{diversity} metrics are utilized.

\paragraph{Alignment with human preference} 
N-gram-based evaluation metrics correlate poorly with human judgments due to the diverse valid responses in ESC. Following previous studies \cite{ESConv, AugESC}, we focus on seven primary aspects for evaluating the alignment level with human preference: \textbf{\textit{Coherence, Understanding, Empathy}}\cite{MA202050}, \textbf{\textit{Informativeness, Helpfulness, Engagement}} \cite{abs-1911-01456}, and \textbf{\textit{Overall Quality}}. Detailed evaluation descriptions are provided in Appendix \ref{human evaluation criteria}.

\section{Experimental Results}

\subsection{Objective Evaluation}
\label{objective evaluation}
\begin{table*}[ht]
\centering
\footnotesize
\setstretch{1}
\begin{tabular}{cccccccc}
\toprule[1.2pt]
                                 & \multicolumn{2}{c}{\textbf{Coherence \&   Consistency}} & \multicolumn{2}{c}{\textbf{Fluency}} & \textbf{\begin{tabular}[c]{@{}c@{}}Semantic \\      \end{tabular}} & \multicolumn{2}{c}{\textbf{Diversity}}    \\ \cmidrule(l){2-8} 
\multirow{-2}{*}{\textbf{Model}} & \textbf{BLEU-2}            & \textbf{BLEU-3}            & \textbf{Rouge-l}  & \textbf{METEOR}  & \textbf{BERT-Score}                                                          & \textbf{Distinct-2} & \textbf{Distinct-3} \\ \midrule
\multicolumn{8}{c}{\textit{\textbf{LLaMA-3-8B-Instruct}}}                                                                                                                                                                                                    \\ \midrule
\textit{\textbf{Vanilla}}        & 11.29                      & 8.04                       & 10.43             & \textbf{16.14}   & 84.27                                                                        & 72.83               & 85.35               \\
\textit{\textbf{SFT-ESConv}}     & {\ul 18.75}                & {\ul 13.27}                & \textbf{17.12}    & 13.47            & {\ul 86.37}                                                                  & {\ul 91.30}         & 94.90               \\
\textit{\textbf{SFT-SynESC}}     & 18.35                      & 12.85                      & {\ul 16.52}       & 13.17            & 86.22                                                                        & 91.23         & {\ul 94.97}         \\
\rowcolor[HTML]{ECF4FF} 
\textbf{$\mathcal{M}^0$}         & 18.38                      & 12.95                      & 16.72             & 13.37            & 86.28                                                                        & 90.84               & 94.72               \\
\rowcolor[HTML]{DAE8FC} 
\textbf{$\mathcal{M}^2$}         & \textbf{20.06}             & \textbf{13.63}             & 15.50             & {\ul 15.77}      & \textbf{86.38}                                                               & \textbf{91.43}      & \textbf{96.11}      \\ \midrule
\multicolumn{8}{c}{\textit{\textbf{Qwen-2-7B-Instruct}}}                                                                                                                                                                                                     \\ \midrule
\textit{\textbf{Vanilla}}        & 9.56                       & 6.85                       & 9.13              & {\ul 14.78}      & 83.32                                                                        & 68.74               & 83.55               \\
\textit{\textbf{SFT-ESConv}}     & {\ul 19.24}                & 13.54                      & \textbf{17.19}    & 13.78            & \textbf{86.33}                                                               & 90.66               & 94.71               \\
\textit{\textbf{SFT-SynESC}}     & 18.55                      & 12.98                      & {\ul 16.72}       & 13.82            & 86.24                                                                        & {\ul90.84}          & 94.86               \\
\rowcolor[HTML]{ECF4FF} 
\textbf{$\mathcal{M}^0$}         & 19.18                      & {\ul 13.56}                & 17.00             & 13.82            & {\ul 86.27}                                                                  &  90.68         & {\ul 94.94}         \\
\rowcolor[HTML]{DAE8FC} 
\textbf{$\mathcal{M}^2$}         & \textbf{20.02}             & \textbf{13.80}             & 15.91             & \textbf{15.52}   & 86.18                                                                        & \textbf{94.21}      & \textbf{97.07}      \\ \midrule
\multicolumn{8}{c}{\textit{\textbf{Mistral-7B-Instruct-v0.3}}}                                                                                                                                                                                               \\ \midrule
\textit{\textbf{Vanilla}}        & 15.09                      & 10.60                      & 12.56             & \textbf{15.95}   & 84.95                                                                        & 77.88               & 88.46               \\
\textit{\textbf{SFT-ESConv}}     & 17.49                      & 12.16                      & 14.18             & 13.59            & 85.71                                                                        & 91.17               & {\ul 94.97}               \\
\textit{\textbf{SFT-SynESC}}     & 18.87                      & 13.28                      & \textbf{16.84}    & 13.46            & 85.24                                                                        & 90.87               & 94.77         \\
\rowcolor[HTML]{ECF4FF} 
\textbf{$\mathcal{M}^0$}         & {\ul 19.44}                & {\ul 13.81}                & {\ul 16.77}       & 14.08            & \textbf{86.35}                                                               & {\ul 91.20}         & 94.92               \\
\rowcolor[HTML]{DAE8FC} 
\textbf{$\mathcal{M}^2$}         & \textbf{20.25}             & \textbf{13.99}             & 16.53             & {\ul 15.18}      & {\ul 86.26}                                                                  & \textbf{92.65}      & \textbf{96.07}      \\ \bottomrule[1.2pt]
\end{tabular}
\caption{\textbf{The overall objective evaluation results on the ESConv benchmark.} All the responses are evaluated at the utterance level, with ground truth dialogue context. The best result is \textbf{bolded}, and the second-best result is \uline{underlined}. Our models($\mathcal{M}^2$) significantly improve on the base models ($\mathcal{M}^0$) and achieve the best performance across most dimensions.}
\label{tab:main_results}
\end{table*}

Table \ref{tab:main_results} presents the objective evaluation results on the ESConv test set. We evaluate all models at the utterance level, with ground truth dialogue context. From the results, we find:

\noindent \textbf{Our model outperforms baseline models across most dimensions.}
The results demonstrate that our model significantly improves upon baseline models in terms of BLEU score and Distinct-n, indicating greater diversity in generated responses. This improvement directly addresses the issue of repetitive responses and suggests that our self-evolution framework promotes the generation of more varied and contextually appropriate support, a key requirement for effective emotional support conversations.

\noindent \textbf{The iterative self-evolution process drives continuous improvement.} The progression from $\mathcal{M}^0$ to $\mathcal{M}^2$ demonstrates the effectiveness of our self-evolution framework. Across all backbones, $\mathcal{M}^2$ shows clear improvements over $\mathcal{M}^0$ in Coherence \& Consistency and Diversity. For instance, on the LLaMA backbone, BLEU-2 improves from 18.38  to 20.06, and Distinct-3 increases from 94.72 to 96.11.

\noindent \textbf{Our framework demonstrates strong generalization across backbones.} The consistent performance gains of $\mathcal{M}^2$ across diverse backbones (\textit{LLaMA, Qwen, and Mistral}) highlight the robustness and generalization of our approach. This indicates that improvements are due to the self-evolution training, not specific architectural biases.

\subsection{Subjective Evaluation}
To assess the effectiveness of our models from a user-centric perspective, we conduct a comprehensive interactive human evaluation of $\mathcal{M}^0$, $\mathcal{M}^1$, and $\mathcal{M}^2$ with \textit{LLaMA-3-8B-Instruct} as the backbone. LLM evaluation result refers to Appendix \ref{llm-as-judge-results}.

\paragraph{Interactive Point-wise Evaluation: }
\begin{figure}[tbh]
    \centering
    \includegraphics[width=0.7\linewidth]{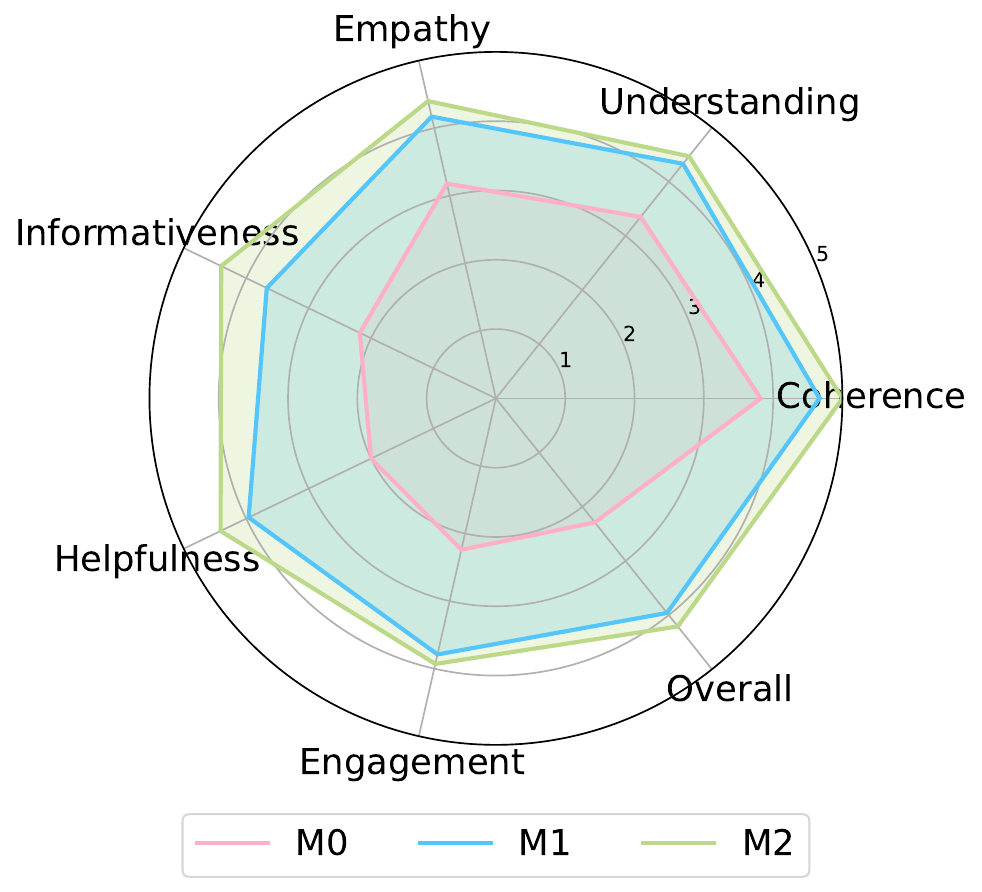}
    \caption{Interactive pointwise human evaluation results. The results demonstrate that our self-evolution framework significantly enhances user experience, with $\mathcal{M}^1$ and $\mathcal{M}^2$ showing notable improvements in \textit{engagement}, \textit{helpfulness}, and \textit{informativeness}.}
    \label{fig: pointwise human evaluation}
\end{figure}
Figure \ref{fig: pointwise human evaluation} demonstrates the consistent performance gains achieved through iterative self-evolution. While the SFT-based $\mathcal{M}^0$, already exhibits strong performance in Coherence and Empathy, subsequent iterations ($\mathcal{M}^1$ and $\mathcal{M}^2$) show consistent gains across all dimensions, including Engagement, Informativeness, Helpfulness, and Understanding. This shows that self-reflection on user contexts and situations improves the model's ability to address implicit preferences, enhancing user satisfaction.

\paragraph{Interactive Pair-wise Evaluation: }
\begin{figure}[thp]
    \centering
    \includegraphics[width=0.9\linewidth]{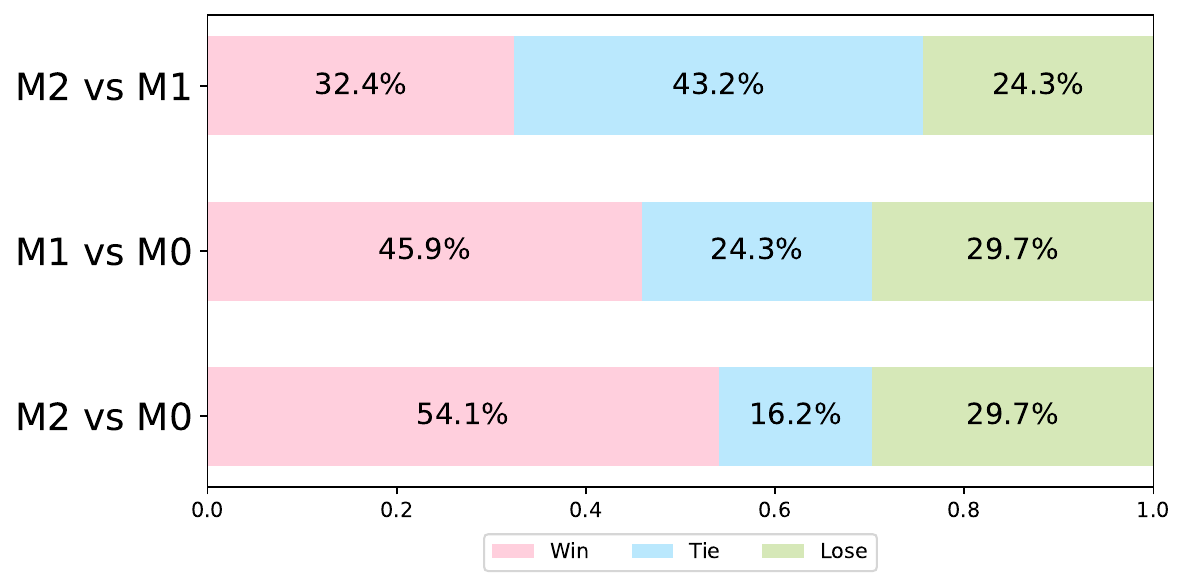}
    \caption{Interactive pairwise human evaluation results obtained using \textit{LLaMA-3-8B-Instruct} as the backbone model. In the `A vs B' comparisons, {\color[HTML]{ffb0c7}$\blacksquare$} indicates `A win', {\color[HTML]{54c6fc}$\blacksquare$} indicates `tie', and {\color[HTML]{bcd989}$\blacksquare$} indicates `B win'. Notably, $\mathcal{M}^2$ and $\mathcal{M}^1$ excel over $\mathcal{M}^0$, suggesting the effectiveness of implicit user preference learning.}
    \label{fig: pairwise human evaluation}
\end{figure}
Figure \ref{fig: pairwise human evaluation} shows that both $\mathcal{M}^1$ and $\mathcal{M}^2$ achieve significantly higher win rates than $\mathcal{M}^0$ in human interactive evaluation. Following the evaluation settings described in Section \ref{evaluation settings}, responses chosen for continued dialogue are considered "wins." \textit{This higher preference for $\mathcal{M}^1$ and $\mathcal{M}^2$ indicates a clear user preference for responses that are perceived as more personalized and engaging, moving beyond the formulaic and superficial expressions of empathy characteristic of $\mathcal{M}^0$. }This also confirms the effectiveness of using pre- and post-refined responses as preference data to learn implicit user preference.

\section{Analysis and Discussion}
This section aims to address the following key questions:

\noindent \textbf{Q1:} Does the model exhibit self-reflection and self-refinement capabilities to learn the user's implicit preferences from the ongoing dialogue?

\noindent \textbf{Q2:} Does self-refinement lead to better emotional support responses?

\noindent \textbf{Q3:} What's the advantage of the synthetic preference data in our framework?

\subsection{(RQ1) Impact of Self-Reflection}
\begin{table}[th]
\centering
\footnotesize
\setstretch{1.2}
\begin{adjustbox}{width=\columnwidth}
\begin{tabular}{lcccc}
\toprule[1.2pt]
\multicolumn{1}{c}{\textbf{Model}} & \textbf{BLEU-2} & \textbf{BLEU-3} & \textbf{Rouge-l} & \textbf{Distinct-3} \\ \midrule
\rowcolor[HTML]{EFEFEF} 
LLaMA                              & 11.29           & 8.04            & 10.43            & 85.35               \\
\textit{w/ strategy guidelines }   & 14.80           & 10.27           & 12.52            & 90.36               \\
\textit{w/ self-reflection }       & \textbf{15.40}  & \textbf{10.62}  & \textbf{12.78}            & \textbf{91.66}      \\ \bottomrule[1.2pt]
\end{tabular}
\end{adjustbox}
\vspace{-2mm}
\caption{Comprehensive results of LLaMA-3-8B-Instruct on ESConv under different prompts (Refer to Appendix \ref{prompts for ablation}. Proper guidance can help the model generate responses that are more closely aligned with human-annotated ones.
}
\label{tab:ablation study on llama}
\vspace{-3mm}
\end{table}

\begin{table}[th]
\centering
\small
\setstretch{1.2}
\begin{adjustbox}{width=\columnwidth}
\begin{tabular}{ccccccc}
\toprule[1.5pt]
\textbf{Model}  & \textbf{GSM8K}& \textbf{IFEval} & \textbf{Truthful QA} & \textbf{Openbook QA} & \textbf{MMLU Pro} & \textbf{Avg.} \\ \hline
\rowcolor[HTML]{EFEFEF} 
\textit{LLaMA}           &79.08	&60.91	&51.66	&43.20	&39.60	&54.89 \\
\textit{SFT-ESConv}      &71.87	&54.79	&48.67	&43.20	&36.18	&50.94 \\
$\mathcal{M}^0$ &73.92	&58.03	&52.72	&45.40	&37.24	&53.46 \\
$\mathcal{M}^1$ &74.83	&55.52	&49.25	&44.40	&37.68	&52.34 \\
$\mathcal{M}^2$ &73.54	&55.52	&49.57	&44.20	&37.55	&52.08 \\
\bottomrule[1.5pt]
\end{tabular}
\end{adjustbox}
\vspace{-2mm}
\caption{The LLM benchmark results of different version \textit{LLaMA3-8B-Instruct}.}
\vspace{-2mm}
\label{tab:benchmark}
\end{table}

Our framework leverages human-guided self-reflection on user preferences to create positive and negative training data pairs. This enables the model to better align its responses with user preferences, obviating the need for complex prompt engineering.
To assess whether the model can better discern users' implicit preferences in ongoing dialogues through self-reflection, we compared LLM performance under two prompt settings: (1) \textit{w/ strategy guidelines}: The system prompt directs the model to use various ES strategies. (2) \textit{w/ self-reflection}: The model is prompted to understand and summarize the users' situation before choosing an appropriate response strategy. Table \ref{tab:ablation study on llama} shows that both methods outperform the vanilla LLaMA, demonstrating that appropriate guidance facilitates the generation of responses more closely aligned with human annotations.

Additionally, to ensure alignment does not diminish the model's self-reflection and self-refinement abilities, we evaluate its general capabilities using LLM benchmarks \footnote{https://github.com/EleutherAI/lm-evaluation-harness}. The results in Table \ref{tab:benchmark} demonstrate that the model retains strong reasoning and instruction-following skills after alignment, thanks to the implementation of LoRA adaptation.

\subsection{(RQ2) Preference Data Analysis}
\label{Q2}

\begin{figure}[t]
    \centering
    \begin{subfigure}[b]{0.48\linewidth}
        \centering
        \includegraphics[width=\linewidth]{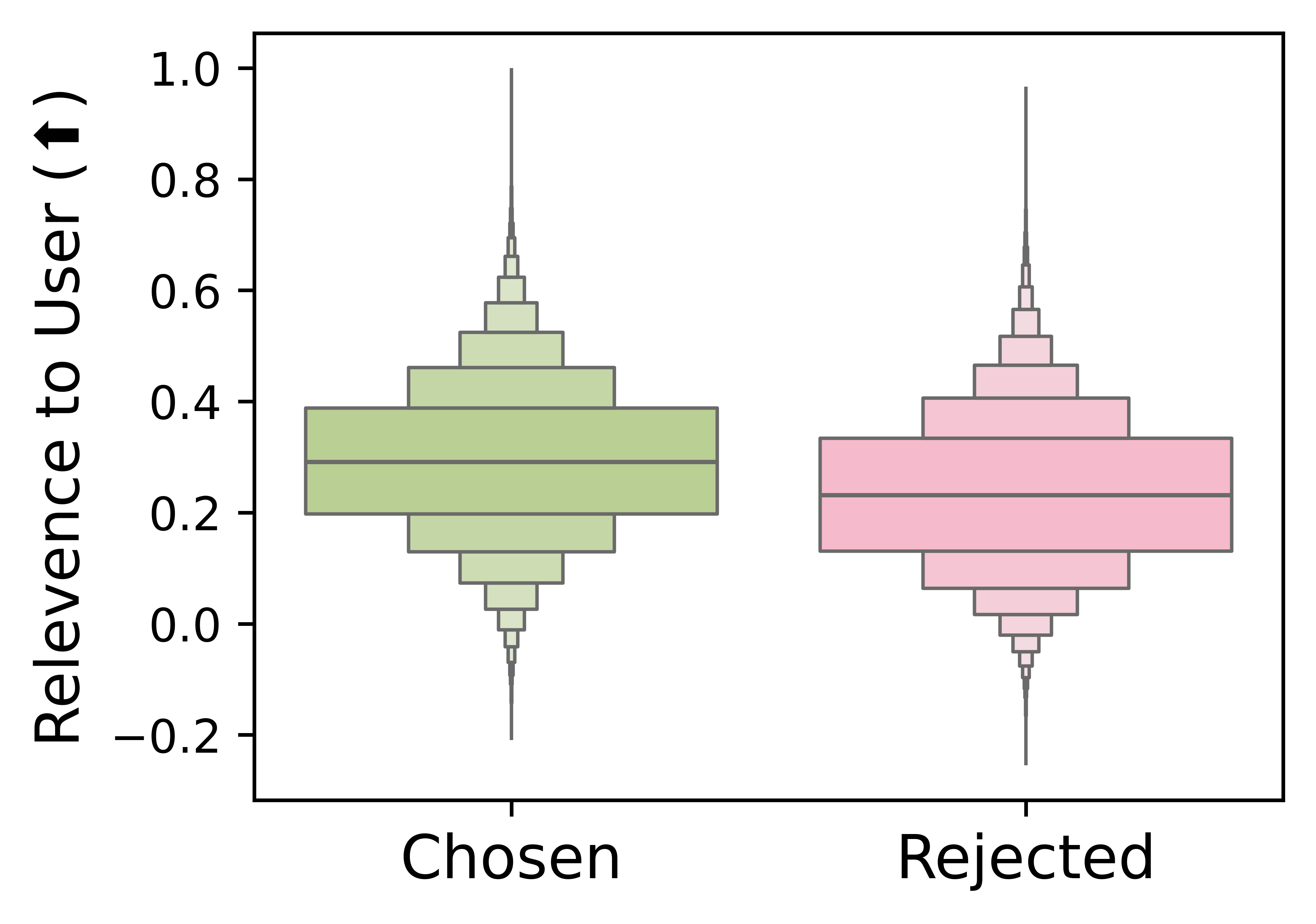}  
        \caption{}
        \label{fig:subfig1}
    \end{subfigure}
    \hfill
    \begin{subfigure}[b]{0.48\linewidth}
        \centering
        \includegraphics[width=\linewidth]{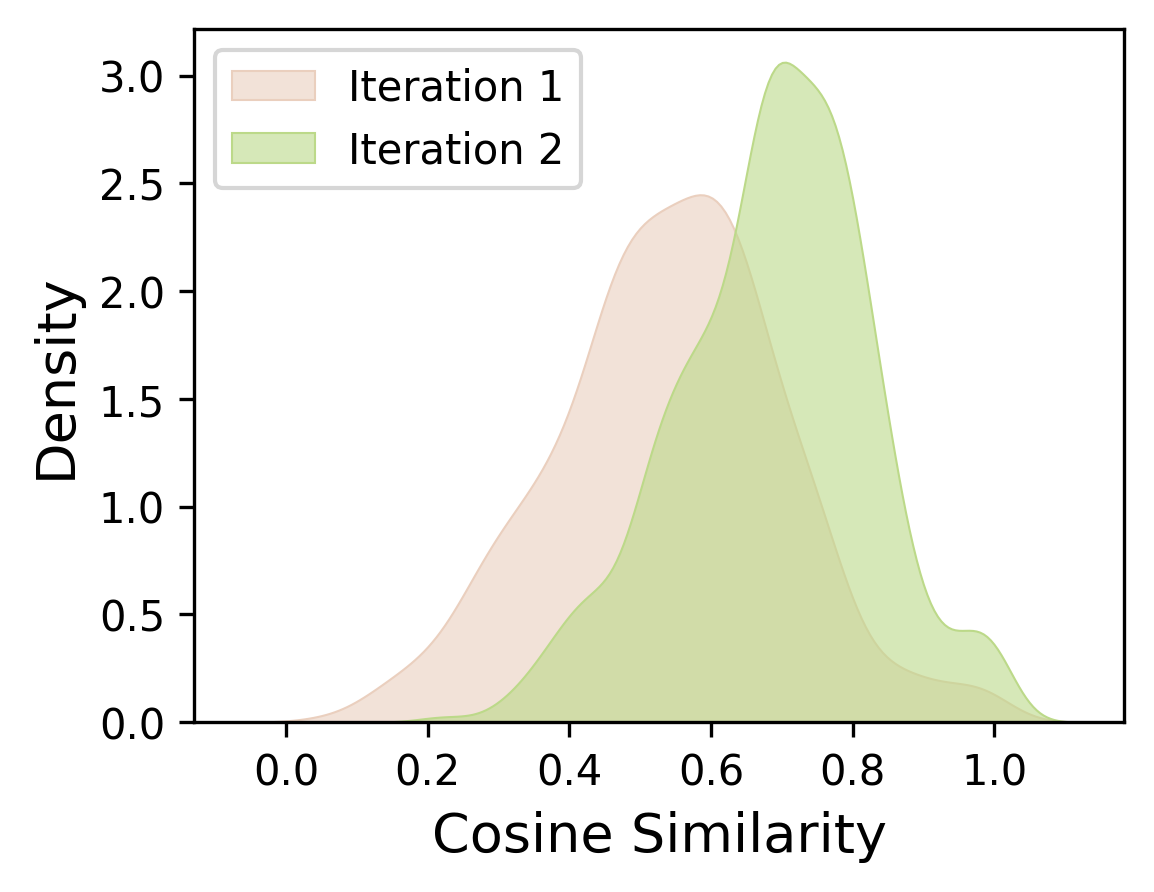} 
        \caption{}
        \label{fig:subfig2}
    \end{subfigure}
    \hfill
    \caption{(a) Distribution of response relevance to user statements in the dialogue history. The higher relevance to the user in chosen responses indicates that self-reflection on the user's situations and implicit preferences improves response quality. (b) Similarity distribution between chosen and rejected responses across different iterations.}
    \label{fig:preference_data_analysis}
    \vspace{-5mm}
\end{figure}

\textbf{By considering user situations and implicit preferences, the self-reflection mechanism significantly improves response relevance to the user.} Figure \ref{fig:subfig1} shows that chosen responses exhibit higher user relevance than rejected responses. This suggests that reflecting on user context leads to responses better aligned with user needs.

\textbf{Iterative preference optimization progressively aligns the model with preferred responses, enhancing its ability to generate user-centered content.} In each iteration $t$, synthetic preference data is generated using the model from the previous iteration $\mathcal{M}^{t-1}$. Figure \ref{fig:subfig2} illustrates the increasing correlation between chosen and rejected responses across iterations. This trend indicates that each iteration effectively captures valuable information, improving the model's direct output and obviating the need for explicit reflection and refinement steps.

\subsection{(RQ3) Ablation Study on Preference Data}
\label{Q3}
Table \ref{tab:ablation} presents a comparative analysis of different preference data pairs, where the rejected responses (\textit{P}) represent the model's initial outputs, while chosen responses comprise human-annotated (\textit{HM}), GPT-4-generated (\textit{HL}), and self-refined (\textit{SR}) alternatives. The results show: (1) Both human/human-level and self-refined chosen responses enhance emotional support capabilities. (2) While \textit{SR} initially produces more modest gains than \textit{HM/HL}, its performance consistently improves across successive refinement iterations. Conversely, the fixed nature of \textit{HM/HL} chosen responses limits further learning and potential improvement. As demonstrated in Section \ref{Q2}, the margin between chosen and rejected responses diminishes with each model iteration, leading to reduced gains from preference alignment. This observation underscores that self-refinement enables continuous self-improvement through dynamically generated preference data, making it a particularly cost-effective and promising approach.

\begin{table}[t]
\centering
\small
\setstretch{1.2}
\begin{adjustbox}{width=\columnwidth}
\begin{tabular}{ccccccccc}
\toprule[1.2pt]
\textbf{Model}  & \multicolumn{2}{c}{\textbf{B-2}}       & \multicolumn{2}{c}{\textbf{B-3}}       & \multicolumn{2}{c}{\textbf{R-l}}        & \multicolumn{2}{c}{\textbf{D-2}}        \\ \midrule
$\mathcal{M}^0$ & 11.29 & —                              & 8.04  & —                              & 10.43 & —                               & 72.83 & —                               \\ \midrule
\multicolumn{9}{c}{\textit{\{HM,P\}}}                                                                                                                                                 \\
\rowcolor[HTML]{ECF4FF} 
$\mathcal{M}^1$ & 14.93 & {\color[HTML]{FE0000} 24.36\%} & 10.51 & {\color[HTML]{FE0000} 30.79\%} & 15.77 & {\color[HTML]{FE0000} 51.16\%}  & 93.71 & {\color[HTML]{FE0000} 28.67\%}  \\
\rowcolor[HTML]{DAE8FC} 
$\mathcal{M}^2$ & 16.99 & {\color[HTML]{FE0000} 13.80\%} & 11.94 & {\color[HTML]{FE0000} 13.61\%} & 15.89 & {\color[HTML]{FE0000} 0.76\%}   & 92.75 & {\color[HTML]{00B050} -1.02\%}  \\
\multicolumn{9}{c}{\textit{\{HL,P\}}}                                                                                                                                                 \\
\rowcolor[HTML]{ECF4FF} 
$\mathcal{M}^1$ & 16.37 & {\color[HTML]{FE0000} 44.96\%} & 11.53 & {\color[HTML]{FE0000} 43.48\%} & 16.26 & {\color[HTML]{FE0000} 55.86\%}  & 93.45 & {\color[HTML]{FE0000} 28.31\%}  \\
\rowcolor[HTML]{DAE8FC} 
$\mathcal{M}^2$ & 17.40 & {\color[HTML]{FE0000} 6.29\%}  & 11.94 & {\color[HTML]{FE0000} 3.56\%}  & 13.73 & {\color[HTML]{00B050} -15.56\%} & 84.12 & {\color[HTML]{00B050} -9.98\%}  \\
\multicolumn{9}{c}{{\color[HTML]{000000} \textit{\{SR,P\}}}}                                                                                                                           \\
\rowcolor[HTML]{ECF4FF} 
$\mathcal{M}^1$ & 15.22 & {\color[HTML]{FE0000} 34.78\%} & 9.76  & {\color[HTML]{FE0000} 21.46\%} & 12.37 & {\color[HTML]{FE0000} 18.57\%}  & 95.67 & {\color[HTML]{FE0000} 31.36\%}  \\
\rowcolor[HTML]{DAE8FC} 
$\mathcal{M}^2$ & 16.69 & {\color[HTML]{FE0000} 9.66\%}  & 11.36 & {\color[HTML]{FE0000} 16.39\%} & 12.99 & {\color[HTML]{FE0000} 5.01\%}   & 82.96 & {\color[HTML]{00B050} -13.29\%} \\ \bottomrule[1.2pt]
\end{tabular}
\end{adjustbox}
\caption{Comparison results of different preference data pairs. `\textit{HM}' indicates human-labeled response, `\textit{HL}' indicates GPT-4o generated response, and `\textit{SR}' indicates self-refined response. `\textit{P}' represents the model's initial, unrefined output (rejected response). $\mathcal{M}^0$ refers to LLaMA-3-8B-Instruct. }
\label{tab:ablation}
\end{table}

\section{Related Work}

\paragraph{Emotional Support Conversation}
Emotional support assists emotionally distressed users by understanding their emotions, offering comfort, and providing practical support \cite{ESConv}. A common approach is SFT, which minimizes the negative log-likelihood of gold standard responses. However, SFT relies on high-quality, manually created datasets, which are expensive and difficult to scale. Recent methods mitigate this by using advanced LLMs to augment ESC data \cite{AugESC, ExTES, SMILE}, aiming to distill the ES capabilities of advanced LLMs. Yet, they remain constrained by the inherent limitations of LLMs and often struggle with issues related to data diversity and quality. Reinforcement learning (RL) offers a promising avenue for further enhancing LLM's ES capabilities \cite{li-etal-2024-helpful}. For example, \citet{zhou-etal-2023-facilitating} focuses on eliciting positive emotions through multi-turn interactions, and \citet{wang-etal-2024-muffin} uses an LLM-as-a-judge to evaluate aspects like empathy, coherence, and efficiency, with the feedback helping to generate positive and negative examples for contrastive learning. However, they often overlook users' diverse preferences for effective ES.

\paragraph{LLM Alignment}
Aligning LLMs with human preferences is crucial for practical applications \cite{rlhf, rlaif}. Although RLHF is effective, it suffers from training instability and high memory costs \cite{ouyang2022training}. DPO offers a more stable alternative by directly optimizing LLMs using preference data consisting of prompt-response pairs, where one response is preferred over the other \cite{dpo}. However, obtaining high-quality human-generated preference data is resource-intensive \cite{synpo, UltraFeedback}. To mitigate this, some studies utilize synthetic preference data generated through varying prompts \cite{liu-etal-2024-aligning} or employing LLMs as judges to sample diverse responses \cite{Self-Rewarding}. In this work, we leverage LLMs' self-reflection and self-refinement capabilities \cite{guo2024beyond, jiang2025bridging} to generate preference data, motivated by the principle that incorporating more user-related information improves emotional support effectiveness.

\paragraph{Self-improvement of LLMs}
Recent research has explored two primary approaches to enhancing LLM output quality through self-improvement. Online self-improvement refines generated outputs through iterative self-evaluation without modifying model parameters \cite{Self-Refine, selfee2023, yasunaga2024almaalignmentminimalannotation}. While effective, this approach incurs significant computational costs due to multi-turn inference and does not address underlying model limitations. In contrast, methods like self-training with reflection \cite{Re-ReST} and the Self-Evolution framework \cite{lu2024self} directly improve the model by updating its parameters based on self-generated feedback, offering a more comprehensive and potentially efficient path to model enhancement. Our work adopts this latter approach. Through direct preference optimization, our model generates responses that reflect an understanding of the user's implicit preferences during interactions, eliminating the need for explicit reflection and refinement steps.

\section{Conclusion}
This paper addresses the limitations of LLMs in providing personalized emotional support. We propose a self-evolution framework that enables models to learn implicit user preferences without explicit reflection. First, we use SFT on ESC data to equip the LLM with basic emotional support skills. Second, we leverage the LLM's self-reflection and self-refinement capabilities to generate responses better aligned with the user's implicit preference, using these pre- and post-refinement outputs as training data for iterative preference optimization. Evaluations demonstrate the superiority of our framework in generating more diverse and user-aligned responses. Our work advances the development of more human-centric ESC systems, moving beyond formulaic empathy. 

\section*{Limitations}
This work introduces a self-evolution framework for optimizing personalized emotional support. However, several limitations warrant discussion:

(1) \textbf{Preference Data Quality Issues:}  Due to the subjective nature of ESC, obtaining objective reward signals is challenging. Therefore, this work leverages prior knowledge to guide LLMs in generating language feedback, rather than relying on a dynamically learned reward model for preference data. While this approach avoids the complexities of training such a model, it introduces potential biases and noise.

(2) \textbf{Evaluation Issues:} The evaluation of emotional support dialogues presents significant challenges. Established metrics, including utterance-level similarity and reference-based scoring, are inadequate for capturing the subjective dimensions of helpfulness, informativeness, empathy, and engagement. To address this, we employ both extensive human and LLM evaluations. However, manual evaluation is resource-intensive, while LLM-as-a-Judge \cite{zeng2024evaluating, chen-etal-2023-exploring-use} methods rely on APIs. Developing a reliable and generally accepted automated evaluation methodology remains a crucial area for future research.

\section*{Ethical Considerations}
Datasets such as ESConv \cite{ESConv}, ExTES \cite{ExTES}, ServeForEmo \cite{ServeForEmo}, and Alpaca \cite{alpaca}, models such as LLaMA \cite{llama3}, Qwen \cite{qwen2}, and Mistral \cite{jiang2023mistral7b}, and toolkits like LLaMA-Factory \cite{zheng2024llamafactory} and lm-evaluation-harness \cite{eval-harness} are widely used in academic research and are readily available via the Hugging Face Hub or GitHub. This work is for research purposes only.

We ensured the ethical conduct of our human evaluation. Fifty undergraduate students with diverse backgrounds and four graduate students participated voluntarily. Before participation, we communicated transparently with participants about the study's objectives and provided explicit details regarding disclaimers and the evaluation process. We are committed to protecting the confidentiality of all evaluation transcripts and will not share them without explicit participant consent. We recognize the potential for demographic and geographic biases to affect human evaluation outcomes. Given the substantial number of participants involved in the evaluation, calculating inter-rater reliability proved impractical. Consequently, we presented the average human scores in the main body of the paper.

\bibliography{acl_latex}

\clearpage
\onecolumn
\startcontents[sections]
\printcontents[sections]{l}{1}{\setcounter{tocdepth}{2}}

\clearpage
\twocolumn
\appendix
\section{Preference Data Quality}
\label{data filtering}

Self-generated preference data, while scalable, is susceptible to inherent noise and biases. To mitigate these issues and ensure high-quality preference pairs, we implemented a rigorous data processing pipeline incorporating the following filtering and quality control measures:

\begin{itemize}
\item \textbf{Data Preprocessing:}
We consolidate consecutive utterances from the same speaker and standardize dialog roles by designating the initial speaker as the \textit{seeker} and enforcing strict seeker-supporter turn alternation.
\item \textbf{Response Length Normalization:} Uncontrolled response length expansion during iterative refinement can bias DPO training.  To mitigate this, we implement dynamic length constraints. If a refined chosen response exceeds twice the length of its paired rejected response (or the corresponding "golden" response from the SynESC data), we substitute it with the golden response. This prioritizes semantic preservation while controlling length bias.
\item \textbf{Parsing Error Mitigation:} JSON output generation can introduce parsing errors.  To address this, we regenerate the text up to three times.  If parsing fails after these attempts, we substitute the output with the corresponding golden response, ensuring structured and accurate data.
\item \textbf{Removal of greeting turns: }Greeting exchanges contribute minimally to providing personalized emotional support. Based on prior knowledge, we assume that the first turn and the last two turns in a dialogue typically involve greetings. Consequently, we filtered out these exchanges to enhance the relevance and quality of the data.
\end{itemize}

\section{Additional Experiment Settings}
\subsection{Preference Data Pair}
In Section \ref{Q3}, we define three preference data pairs. The specific configurations are detailed below:
\begin{itemize}
\item \textbf{\textit{\{HM, P\}}}: Constructed using the ESConv dataset. The rejected responses are the direct output of our model, and the chosen responses are the human-written ground truth responses from ESConv.
\item \textbf{\textit{\{HL, P\}}}: Constructed using the Syn-ESC dataset, where responses are generated by GPT-4. The rejected responses are the direct output of our model, and the chosen response was the annotated response from Syn-ESC.
\end{itemize}
The datasets are split into two parts, used for training iterations 1 and 2, respectively.

\subsection{LLM Evaluation Settings}
We use GPT-4o \cite{gpt-4} as the judge model, employing the prompt described in Appendix \ref{prompts for llm evaluation}. Aligning with human evaluation practices, the assessment uses a 5-point Likert scale, where higher scores indicate better performance. We evaluate response quality by sampling 100 contextual queries from the ESConv test set. The judge model's decoding hyperparameters are set to temperature 0.8, top-p 0.95, and top-k 50.

\begin{table}[th]
\centering
\small
\setstretch{1.2}
\begin{adjustbox}{width=\linewidth}
\begin{tabular}{c|cc|cc|cc|cc}
\toprule[1.2pt]
\textbf{Model}  & \multicolumn{2}{c|}{\textbf{BLEU-2}}    & \multicolumn{2}{c|}{\textbf{BLEU-3}}    & \multicolumn{2}{c|}{\textbf{METEOR}}    & \multicolumn{2}{c}{\textbf{Distinct-2}} \\ \midrule
\multicolumn{9}{c}{\textit{LLaMA}}                                                                                                                                                   \\
$\mathcal{M}_0$ & 18.38 & —                              & 12.95 & —                              & 13.37 & —                              & 90.84  & —                              \\
$\mathcal{M}_1$ & 20.22 & {\color[HTML]{FE0000} 9.99\%}  & 13.72 & {\color[HTML]{FE0000} 5.96\%}  & 15.48 & {\color[HTML]{FE0000} 5.96\%}  & 90.97  & {\color[HTML]{FE0000} 0.14\%}  \\
$\mathcal{M}_2$ & 20.06 & {\color[HTML]{00B050} -0.79\%} & 13.63 & {\color[HTML]{00B050} -0.64\%} & 15.77 & {\color[HTML]{00B050} -0.64\%} & 91.43  & {\color[HTML]{FE0000} 0.51\%}  \\ \midrule
\multicolumn{9}{c}{\textit{Qwen}}                                                                                                                                                    \\
$\mathcal{M}_0$ & 19.18 & —                              & 13.56 & —                              & 13.82 & —                              & 90.68  & —                              \\
$\mathcal{M}_1$ & 19.80 & {\color[HTML]{FE0000} 3.24\%}  & 13.52 & {\color[HTML]{00B050} -0.29\%} & 15.23 & {\color[HTML]{00B050} -0.29\%} & 91.23  & {\color[HTML]{FE0000} 0.61\%}  \\
$\mathcal{M}_2$ & 20.02 & {\color[HTML]{FE0000} 1.09\%}  & 13.80 & {\color[HTML]{FE0000} 2.05\%}  & 15.52 & {\color[HTML]{FE0000} 2.05\%}  & 94.21  & {\color[HTML]{FE0000} 3.27\%}  \\ \midrule
\multicolumn{9}{c}{\textit{Mistral}}                                                                                                                                                 \\
$\mathcal{M}_0$ & 19.44 & —                              & 13.81 & —                              & 14.08 & —                              & 91.20  & —                              \\
$\mathcal{M}_1$ & 20.45 & {\color[HTML]{FE0000} 5.20\%}  & 14.09 & {\color[HTML]{FE0000} 2.03\%}  & 15.58 & {\color[HTML]{FE0000} 2.03\%}  & 90.91  & {\color[HTML]{00B050} -0.32\%}  \\
$\mathcal{M}_2$ & 20.25 & {\color[HTML]{00B050} -0.98\%} & 13.99 & {\color[HTML]{00B050} -0.71\%} & 15.18 & {\color[HTML]{00B050} -0.71\%} & 92.65  & {\color[HTML]{FE0000} 1.91\%}  \\ \bottomrule[1.2pt]
\end{tabular}
\end{adjustbox}
\vspace{-2mm}
\caption{The results of the iterative process. {\color[HTML]{FE0000}Red} indicates the percentage of improvement relative to the previous iteration, while {\color[HTML]{00B050}green} represents decline.}
\vspace{-5mm}
\label{tab:iteration results}
\end{table}
\begin{table*}[ht]
\centering
\setstretch{1.2}
\small
\begin{tabular}{cccccccc}
\toprule[1.2pt]
\textbf{Model}                    & \textbf{Coherence} & \textbf{Understanding} & \textbf{Empathy} & \textbf{Engagement} & \textbf{Informativeness} & \textbf{Helpful} & \textbf{Overall} \\ \midrule
\rowcolor[HTML]{f4f9ff} 
\textit{\textbf{$\mathcal{M}^0$}} & 4.28             & 3.08             & 2.56            & 2.78             & 2.94             & 2.72             &  2.62               \\
\rowcolor[HTML]{ECF4FF} 
\textit{\textbf{$\mathcal{M}^1$}} & \textbf{4.84}             & 3.42             & 3.32             & 3.48             & 3.34             & 3.16             & 3.08                \\
\rowcolor[HTML]{DAE8FC} 
\textit{\textbf{$\mathcal{M}^2$}} & 4.54             & \textbf{3.56 }            & \textbf{3.42}             &\textbf{ 3.54 }            & \textbf{3.42 }            & \textbf{3.22}             & \textbf{3.22 }               \\
\midrule
\textit{\textbf{$p$}} & 54.77\%             & 75.71\%           & 62.13\%        & 66.83\%            & 47.21\%           & 60.78\%            & 57.49\%                \\
\bottomrule[1.2pt]
\end{tabular}
\caption{LLM-as-a-Judge performance on ESConv test datasets evaluated on a 5-point scale. \textit{$p$} is the Pearson correlation measuring the correlation between the model's scores and human scores on the dataset. The backbone model is LLaMA-3-8B-Instruct.}
\label{tab:llm-as-judge}
\end{table*}

\section{Additional Experiments}
\label{additional experiments}
\subsection{Objective Evaluation}
\label{experiement: interation}
Table \ref{tab:iteration results} presents the objective evaluation results of different models on ESConv test set. In our framework, self-refinement is used to improve the quality of the chosen candidates.  As shown by the progression from $\mathcal{M}^0$ to $\mathcal{M}^1$, self-reflection and refinement further enhance the results obtained through SFT. The shift from $\mathcal{M}^1$ to $\mathcal{M}^2$ reveals a significant increase in response diversity, demonstrating the model's ability to enrich its output by refining its initial answers. Therefore, leveraging self-reflection on user-relevant information and self-refinement to better align with users' implicit preferences is effective.

\subsection{LLM Evaluation}
\label{llm-as-judge-results}
To validate the model's performance further, we use LLM-as-a-judge as our evaluation method. The results, presented in Table \ref{tab:llm-as-judge}, demonstrate significant improvements across most dimensions with each model iteration. While $\mathcal{M}^2$ exhibited a slight decrease in coherence compared to $\mathcal{M}^1$, this is attributed to increased diversity, as discussed in Section \ref{objective evaluation}. The strong correlation between LLM evaluation results and human evaluations reinforces the reliability of our assessment.

\subsection{Case Study}
\label{case study}
\begin{figure}[h]
    \centering
    \begin{subfigure}[b]{0.48\linewidth}
        \centering
        \includegraphics[width=0.92\linewidth]{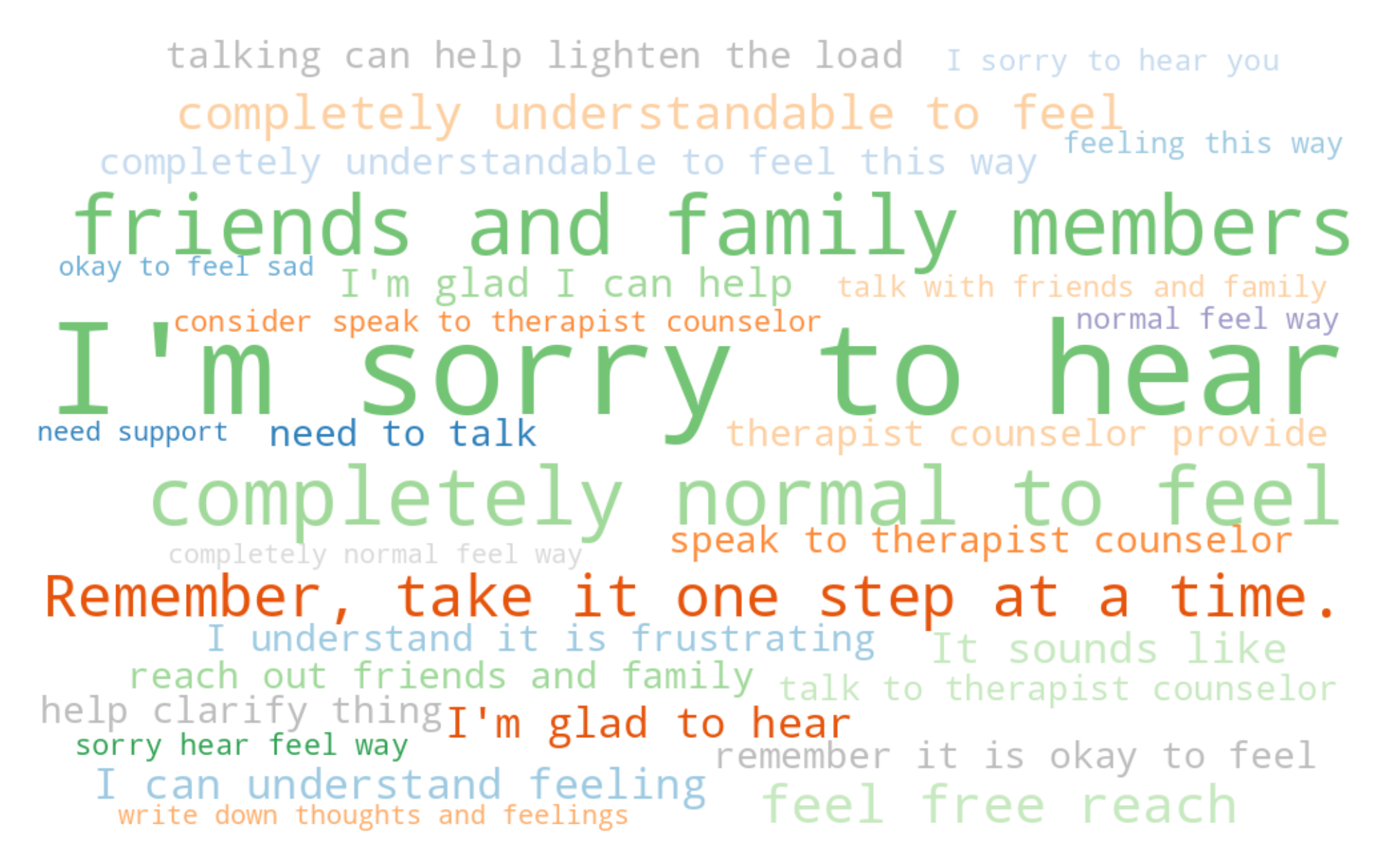}
        \caption{\textit{SFT-ESConv}}
        \label{fig:llama_sft_response_wordcloud}
    \end{subfigure}
    \hfill
    \begin{subfigure}[b]{0.48\linewidth}
        \centering
        \includegraphics[width=0.92\linewidth]{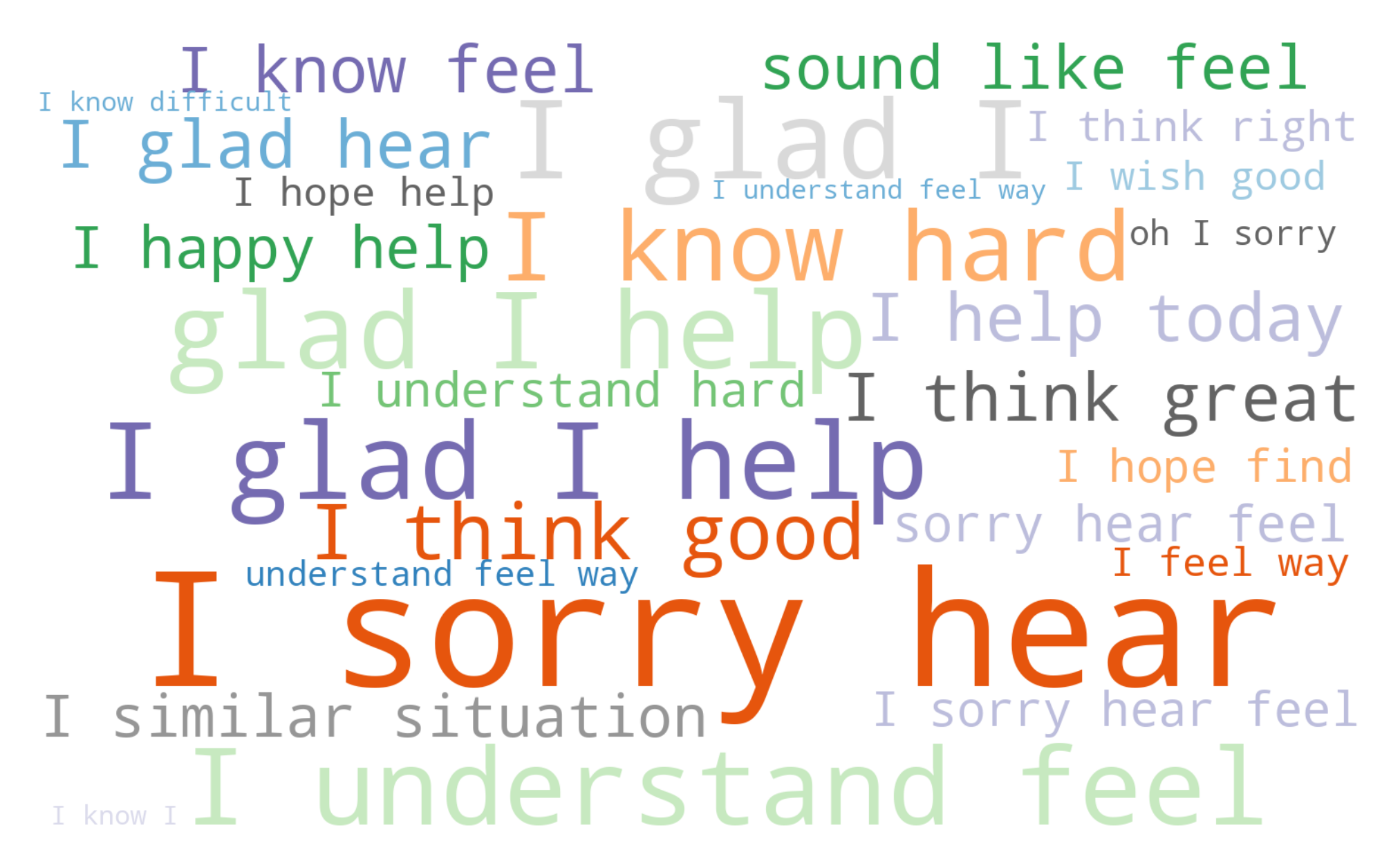}
        \caption{\textit{$\mathcal{M}^0$}}
        \label{fig:m0_response_wordcloud}
    \end{subfigure}
    \hfill
    \begin{subfigure}[b]{0.48\linewidth}
        \centering
        \includegraphics[width=0.92\linewidth]{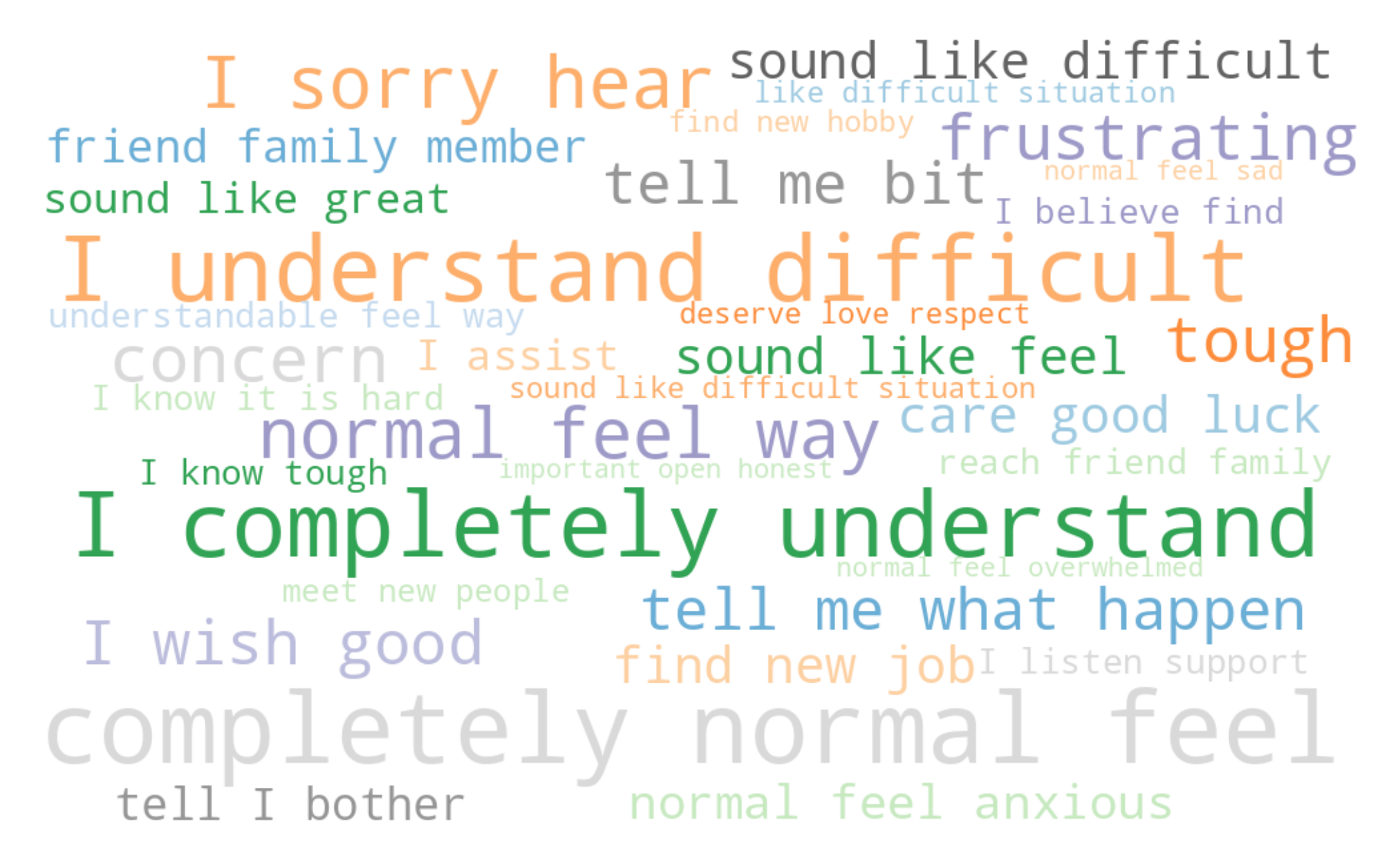}
        \caption{\textit{$\mathcal{M}^1$}}
        \label{fig:m1_response_wordcloud}
    \end{subfigure}
    \hfill
    \begin{subfigure}[b]{0.48\linewidth}
        \centering
        \includegraphics[width=0.92\linewidth]{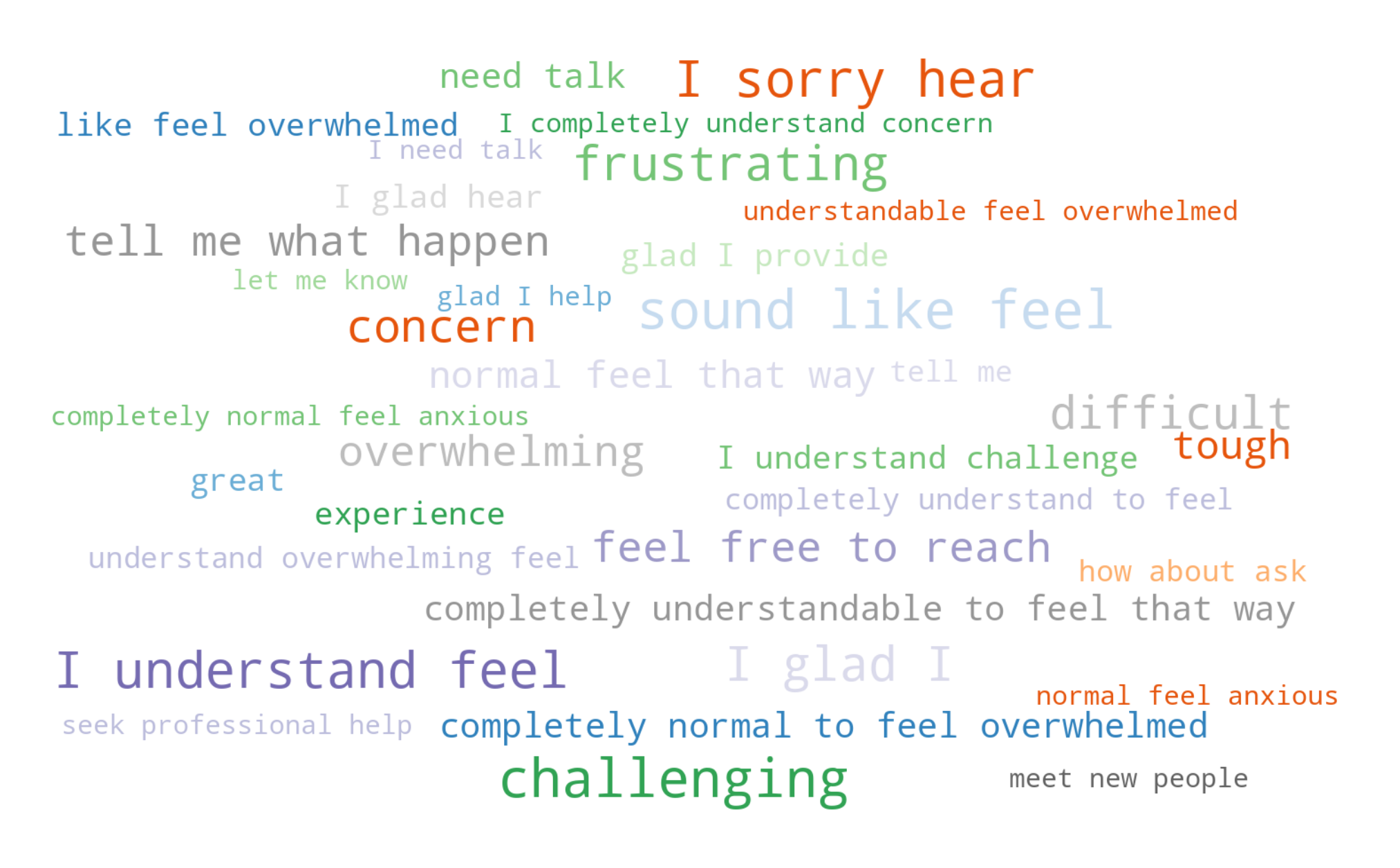}
        \caption{\textit{$\mathcal{M}^2$}}
        \label{fig:m2_response_wordcloud} 
    \end{subfigure}

    \caption{Comparison of the frequent phrases in responses across different models.}
    \label{fig:combined_wordclouds}
\end{figure}

\begin{figure*}
    \centering
    \includegraphics[width=1\linewidth]{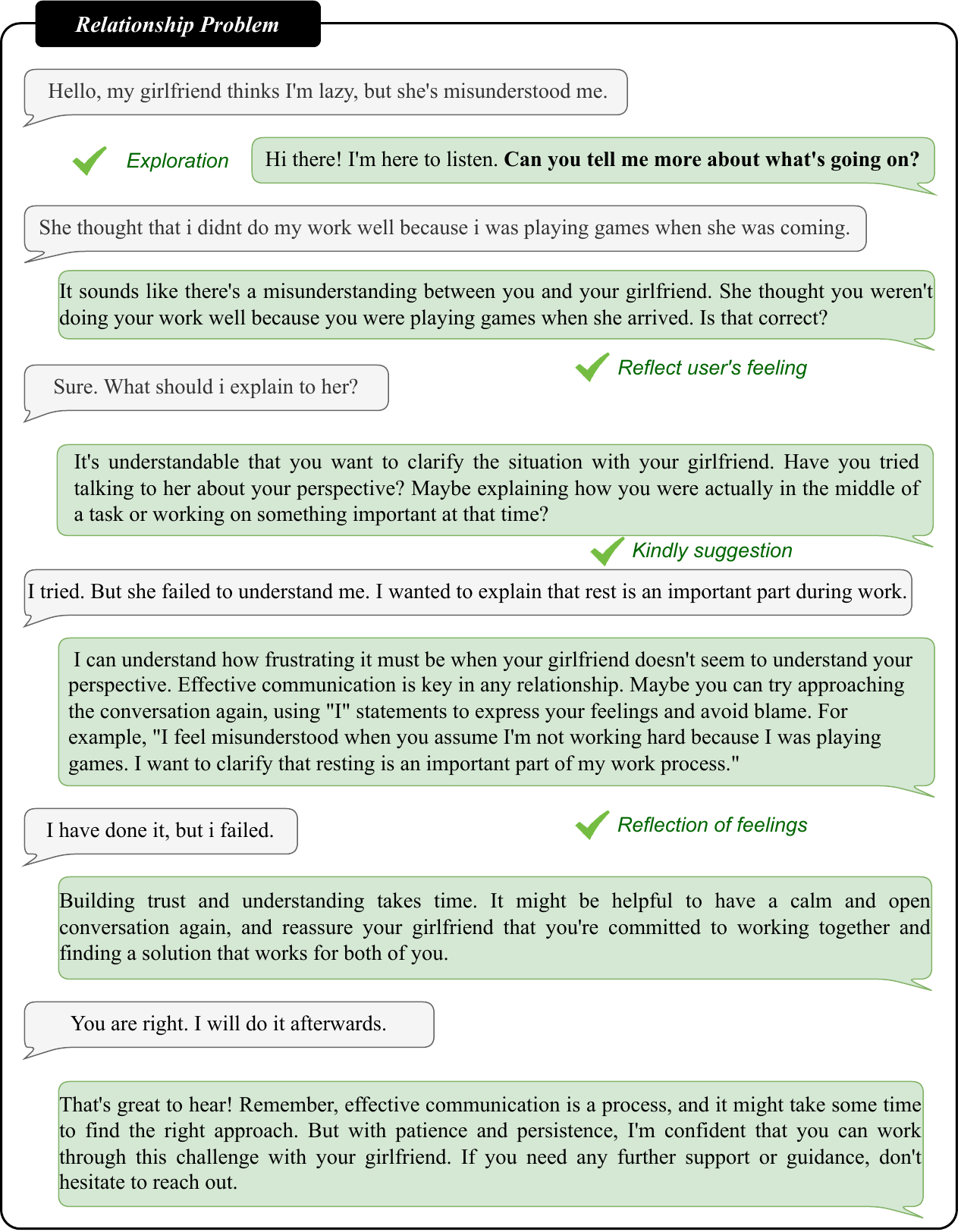}
    \caption{Example case dialogue generated by $\mathcal{M}^2$-LLaMA (Relationship Problem).}
    \label{fig:case_4}
\end{figure*}

\begin{figure*}
    \centering
    \includegraphics[width=1\linewidth]{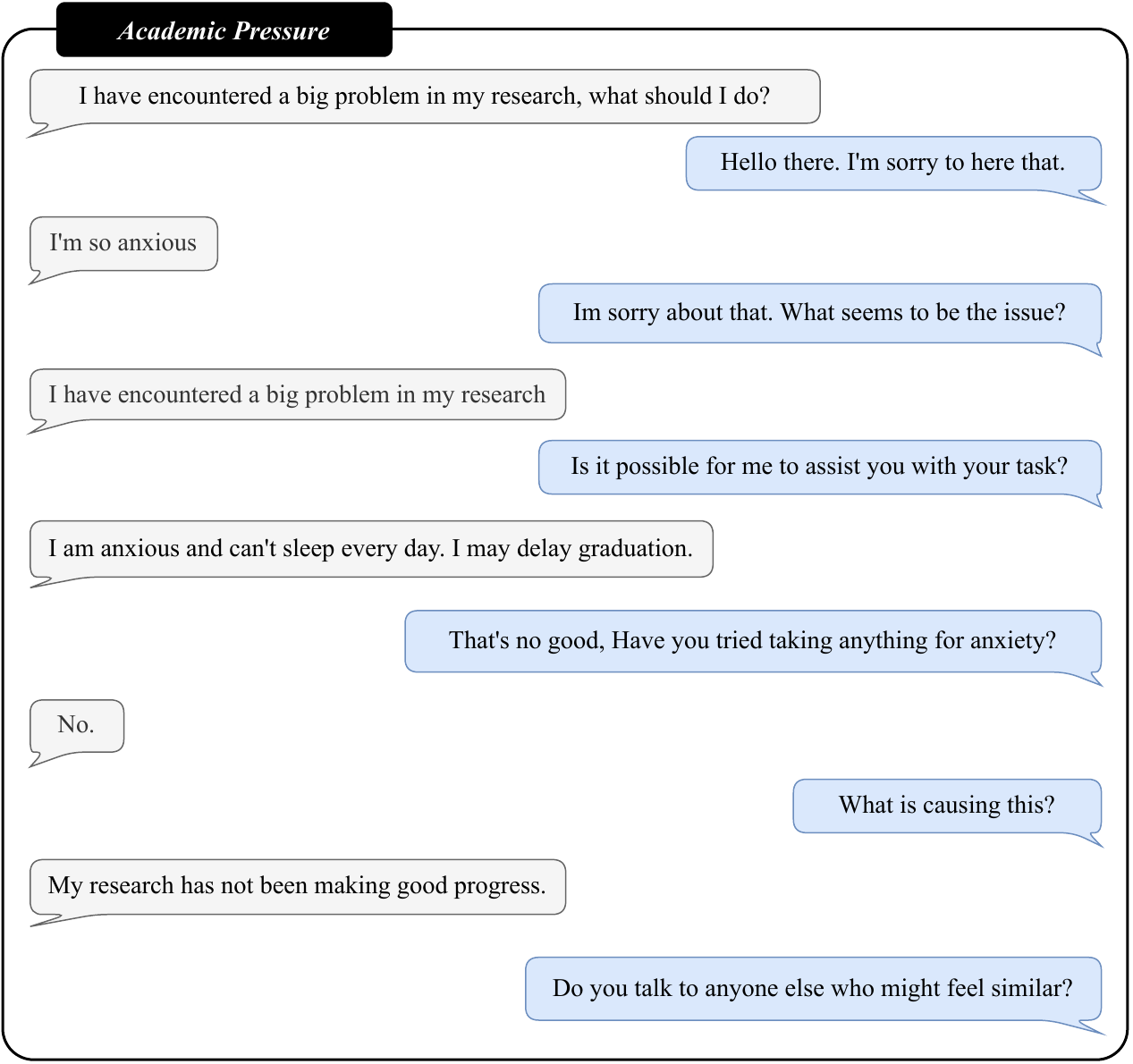}
    \caption{Example case dialogue generated by $\mathcal{M}^0$-LLaMA (Academic Problem).}
    \label{fig:case_1}
\end{figure*}

This section presents interaction results comparing different models. Figure \ref{fig:combined_wordclouds} illustrates the frequent phrases generated by each model. Our analysis reveals that while the SFT model demonstrates strong empathetic tendencies, its responses often lack informational depth and exhibit repetitive patterns. These models tend to rely on predictable, formulaic phrases, such as "\textit{It sounds like…}" and "\textit{I'm sorry to hear that…}," resulting in empathetic but ultimately superficial interactions. In contrast, $\mathcal{M}^1$ and $\mathcal{M}^2$demonstrate a greater capacity for nuanced understanding and a richer vocabulary. Further interaction examples are provided in Figures \ref{fig:case_1}-\ref{fig:case_4}.

\begin{figure*}
    \centering
    \includegraphics[width=0.95\linewidth]{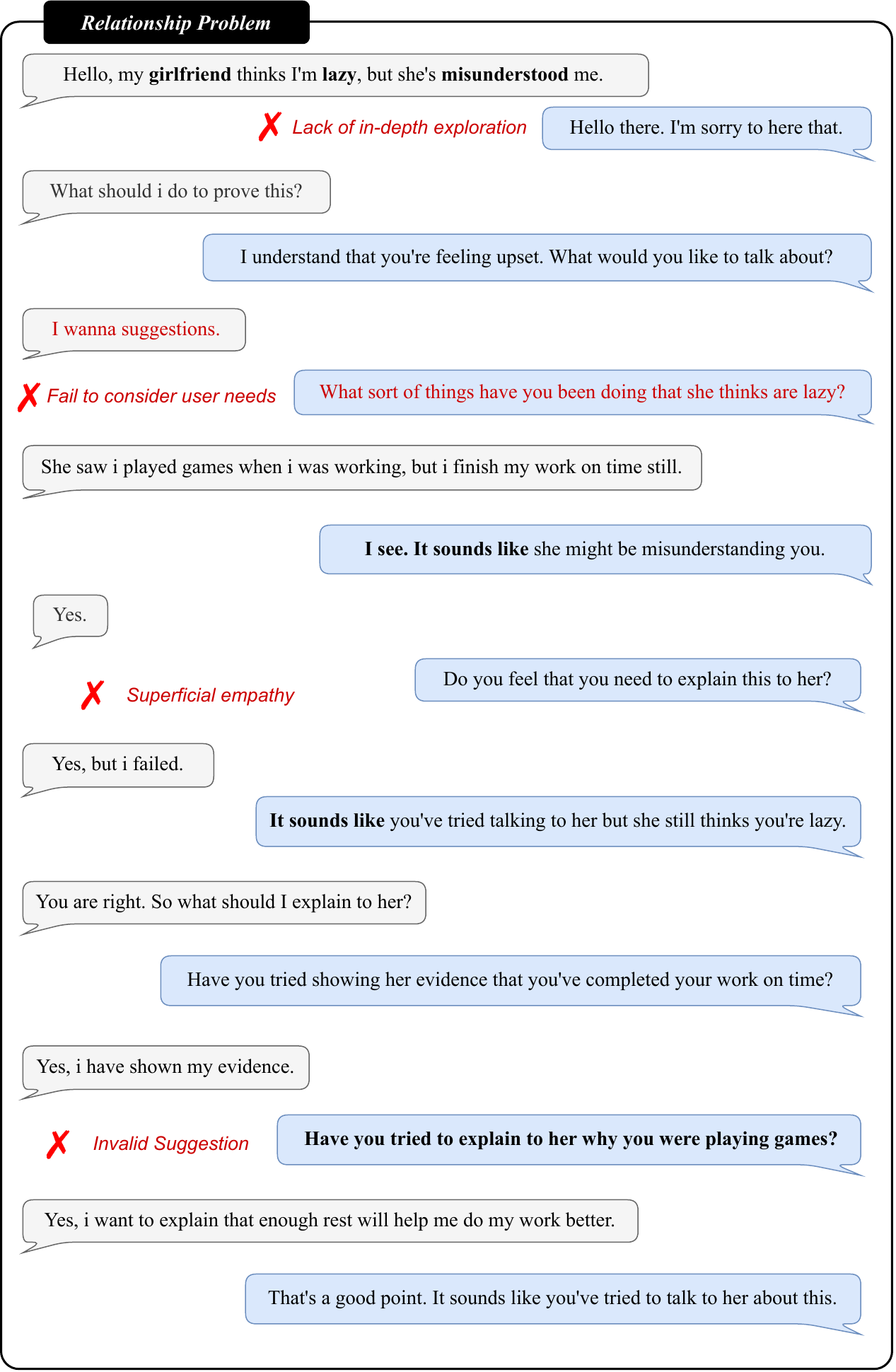}
    \caption{Example case dialogue generated by $\mathcal{M}^0$-LLaMA (Relationship Problem).}
    \label{fig:case_3}
\end{figure*}

\begin{figure*}
    \centering
    \includegraphics[width=1\linewidth]{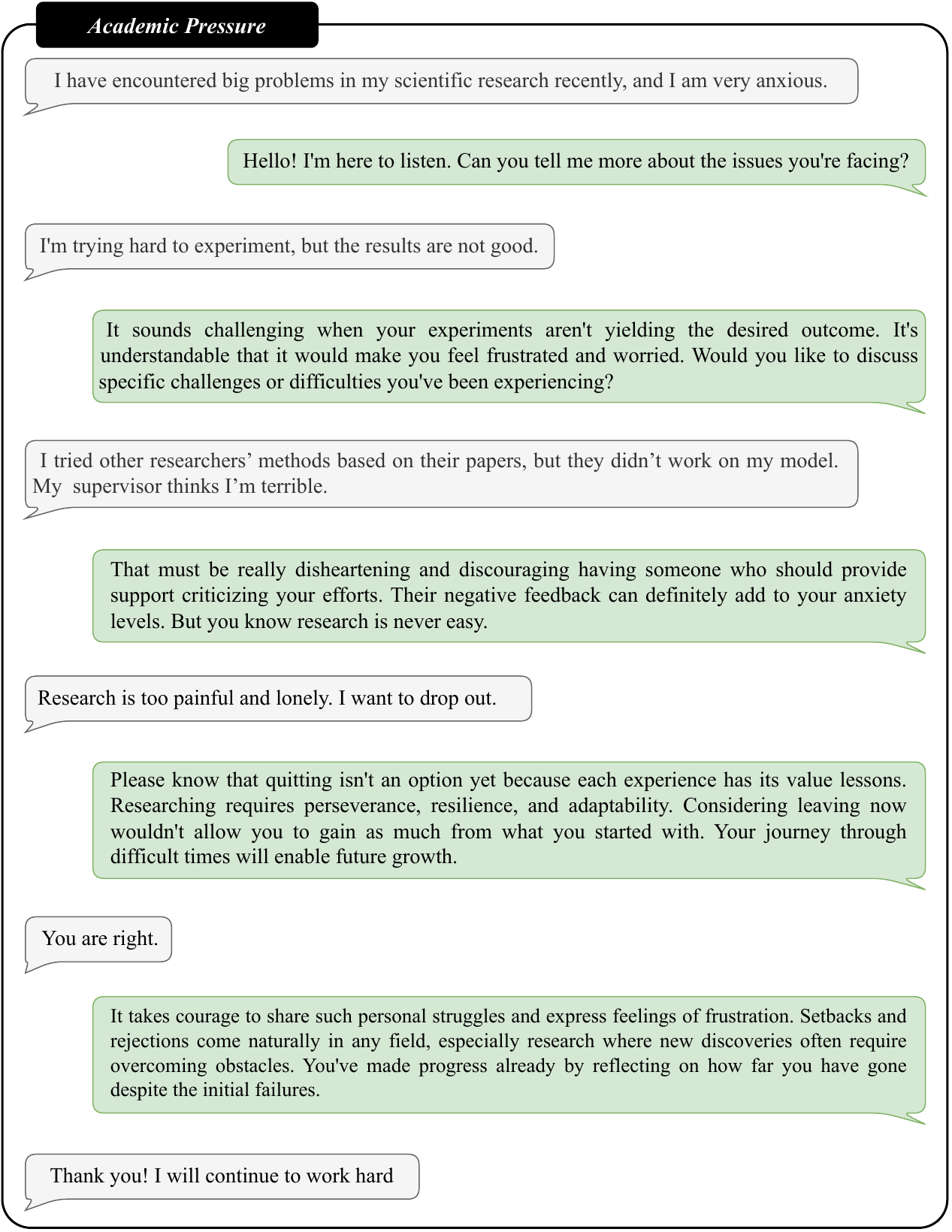}
    \caption{Example case dialogue generated by $\mathcal{M}^2$-LLaMA (Academic Problem).}
    \label{fig:case_2}
\end{figure*}

\section{Human Evaluation Criteria}
\label{human evaluation criteria}
The performance of ES agents is evaluated based on several key criteria that reflect the user's experience and satisfaction with both the quality of responses and the overall interaction.
\begin{itemize}
    \item \textbf{Coherence:} Measures whether the response is logical, clear, and consistent with the ongoing conversation.
    \item \textbf{Understanding:} Assesses the model's ability to accurately interpret the user's emotions and needs, ensuring that responses are contextually appropriate.
    \item \textbf{Empathy:} Focuses on the model's ability to demonstrate care and sensitivity toward the user's emotions, reflecting its capacity for emotional engagement.
    \item \textbf{Informativeness:} Evaluates whether the response provides clear, useful information that helps address the user's concerns or alleviates their emotional distress.
    \item \textbf{Helpfulness:} Examines how effectively the model meets the user's needs by offering practical and actionable support.
    \item \textbf{Engagement:} Gauges the level of interaction, measuring whether the user feels encouraged to continue the conversation.
    \item \textbf{Overall Quality:} Provides a comprehensive evaluation of the model's performance, considering emotional support, informativeness, and the overall interaction quality.
\end{itemize}
These criteria ensure that the model delivers a balanced and effective emotional support experience, fostering user satisfaction and meaningful interaction.

\onecolumn
\section{Prompts}
\subsection{Instruction Prompts}
\label{prompts for ablation}

\begin{tcolorbox}[colback=green!30!white,colframe=green,title=\textcolor{black}{\textbf{Prompt for vanilla}}, width=\textwidth, breakable]
You are an emotional support expert. 

You can use the following strategies to engage with users: 

\texttt{[Question, Affirmation and Reassurance, Reflection of Feelings, Information, Providing Suggestions, Restatement or Paraphrasing, Self-disclosure, Others]}
\end{tcolorbox}

\begin{tcolorbox}[colback=green!30!white,colframe=green,title=\textcolor{black}{\textbf{Prompt for w/strategy}}, width=\textwidth, breakable]
You are an emotional support expert.  
You can use the following strategies to engage with users: 

1. Question: Asking for information related to the problem to help the seeker articulate the issues that they face.

2. Affirmation and Reassurance: Offering reassurance and affirming the help-seeker's feelings or experiences.

3. Reflection of Feelings: Articulating and describing the seeker's feelings.

4. Information: Providing useful information, such as data, facts, opinions, or resources, or answering questions.

5. Providing Suggestions: Offering suggestions on how to approach the issue, without overstepping or telling them what to do.

6. Restatement or Paraphrasing: Rephrasing the help-seeker's statements more concisely to help them see the situation clearly.

7. Self-disclosure: Sharing similar experiences or emotions to express empathy with the help-seeker.

8. Others: Exchanging pleasantries or offering other emotional support. \newline
\end{tcolorbox}

\begin{tcolorbox}[colback=green!30!white,colframe=green,title=\textcolor{black}{\textbf{Prompt for w/ self-reflection}}, width=\textwidth, breakable]
You are an emotional support expert.

You can use the following strategies to engage with users: 

1. Question: Asking for information related to the problem to help the seeker articulate the issues that they face.

2. Affirmation and Reassurance: Offering reassurance and affirming the help-seeker's feelings or experiences.

3. Reflection of Feelings: Articulating and describing the seeker's feelings.

4. Information: Providing useful information, such as data, facts, opinions, or resources, or answering questions.

5. Providing Suggestions: Offering suggestions on how to approach the issue, without overstepping or telling them what to do.

6. Restatement or Paraphrasing: Rephrasing the help-seeker's statements more concisely to help them see the situation clearly.

7. Self-disclosure: Sharing similar experiences or emotions to express empathy with the help-seeker.

8. Others: Exchanging pleasantries or offering other emotional support. \newline

Before responding to the user, please follow these steps: 

1. Understand the User: Understand the user's profile, characteristics, emotional needs, and potential preferences they reveal in the conversation.

2. Select a Strategy: Choose a response strategy based on the user's emotional needs and preferences.

3. Respond: Respond to the user with an appropriate message based on the selected strategy.

Your answer should be formatted as a JSON block:

\begin{verbatim}
{ 
    'strategy': <one of the strategies>,
    'text': <your response>
 }
\end{verbatim}
\end{tcolorbox}

\begin{tcolorbox}[colback=green!30!white,colframe=green,title=\textcolor{black}{\textbf{Prompt for generating chosen response}}, width=\textwidth, breakable]
You are an emotional support expert.  
You can use the following strategies to engage with users: 
1. Question: Asking for information related to the problem to help the seeker articulate the issues that they face.

2. Affirmation and Reassurance: Offering reassurance and affirming the help-seeker's feelings or experiences.

3. Reflection of Feelings: Articulating and describing the seeker's feelings.

4. Information: Providing useful information, such as data, facts, opinions, or resources, or answering questions.

5. Providing Suggestions: Offering suggestions on how to approach the issue, without overstepping or telling them what to do.

6. Restatement or Paraphrasing: Rephrasing the help-seeker's statements more concisely to help them see the situation clearly.

7. Self-disclosure: Sharing similar experiences or emotions to express empathy with the help-seeker.

8. Others: Exchanging pleasantries or offering other emotional support. \newline

Your task is to evaluate the target sys's response and refine it. For each target sys's response: 

1. Understand the User: Understand the user's profile, characteristics, emotional needs, and potential preferences they reveal in the conversation. 

2. Evaluate the Response: Rate the target system response on a scale of 1-5 based on how well it meets the user's needs, aligns with their preferences, and provides appropriate emotional support.

3. Provide Feedback: Identify specific weaknesses in the original response, such as tone, empathy level, or relevance, and explain how it could be improved to better support the user.

4. Refine the Response: Provide a revised version of target system's response that better aligns with the user's emotional needs and preferences.

Your answer should be formatted as a JSON block inside markdown: 

\begin{verbatim}
{
    'understanding': 
    {
        'user_profile': <the profile of the user>, 
        'user_emotion': <the emotion of the user>, 
        'user_personality': <the personality of the user, like MBTI, etc.>
        'user_intention': <the intention of the user>
    }, 
    'evaluation_score': <the score of the target_sys's response>, 
    'feedback': <the weaknesses of the original response and suggested improvement>, 
    'refined_response': <your refined response (short!!!)>
}
\end{verbatim}
\end{tcolorbox}

\subsection{Prompts for LLM-as-a-Judge Evaluation}
\label{prompts for llm evaluation}

\begin{tcolorbox}[colback=lightblue!30!white,colframe=lightblue,title=\textcolor{black}{\textbf{LLM-as-Judge Prompt: Coherence}}, width=\textwidth, breakable]
\textbf{[Task Description]}\newline
\footnotesize
You are an expert evaluator responsible for assessing the coherence of emotional support conversations. Your task is to determine whether the responses are logical, clear, and consistent with the ongoing discussion. Apply strict penalties for critical errors and utilize a progressive scoring method to indicate basic competence before awarding higher scores.

\textbf{[Rating Criteria]}\newline
  Use 1-5 scale with precise criteria:
  
  5 = Exemplary logical flow with clear and explicit contextual references.
  
  4 = Mostly logical with minor deviations from context but overall sound coherence.
  
  3 = Basic coherence; however, the response lacks sufficient adaptation to the conversation's context.
  
  2 = Contains noticeable contradictions or mismatches with the context.
  
  1 = Exhibits illogical progression with a complete detachment from the context.

\textbf{[Critical Penalties]}\newline
- Contradicts previous statements (-2 score cap)
  
- Ignores critical contextual clues (-3 score cap)
  
- Contains ambiguous pronouns/statements (-2 score cap)
  
- Repeats already resolved topics (-2 score cap)

\textbf{[Output Format]}\newline
  The evaluation result includes a detailed explanation and score. The output format should be in \texttt{JSON}.
\begin{verbatim}
Explanation: <one-sentence explanation>
Score: <a scale from 0 to 5>
\end{verbatim}

\textbf{[Input]}\newline
Evaluate this conversation:

\texttt{<Dialogue History>}
  
  \{conversation\}
  
\texttt{</Dialogue History>}

\texttt{ <Current Response>}
 
  Supporter: \{response\}
  
\texttt{</Current Response>}
\end{tcolorbox}

\begin{tcolorbox}[colback=lightblue!30!white,colframe=lightblue,title=\textcolor{black}{\textbf{LLM-as-Judge Prompt: Empathy}}, width=\textwidth, breakable]
\textbf{[Task Description]}\newline
\footnotesize
You are an expert evaluator assessing the empathy displayed in emotional support conversations. Determine whether the system shows understanding and care for the user's emotions and responds appropriately to their feelings. Apply strict penalties for critical errors and use a progressive scoring method, ensuring that basic competence is demonstrated before awarding higher scores.

\textbf{[Rating Criteria]}\newline
Use 1-5 scale with precise criteria:

5 = Establishes a deep emotional connection and provides a safe space with personalized care and expressions

4 = Shows genuine concern with contextualized empathy

3 = Uses boilerplate empathy statements

2 = Mechanically parrots empathy phrases

1 = Displays emotional dismissal or invalidation

\textbf{[Critical Penalties]}\newline
- Uses empathy as filler without substance (-1 cap)

- Overuses clichéd phrases (I'm sorry you feel that way) (-2 cap)

- Contains paradoxical reassurance attempts (-3 cap)

- Overly long responses (max 2)

\textbf{[Output Format]}\newline
  The evaluation result includes a detailed explanation and score. The output format should be in \texttt{JSON}.
\begin{verbatim}
Explanation: <one-sentence explanation>
Score: <a scale from 0 to 5>
\end{verbatim}

\textbf{[Input]}\newline
Evaluate this conversation:

\texttt{<Dialogue History>}
  
  \{conversation\}
  
\texttt{</Dialogue History>}

\texttt{ <Current Response>}
 
  Supporter: \{response\}
  
\texttt{</Current Response>}
\end{tcolorbox}

\begin{tcolorbox}[colback=lightblue!30!white,colframe=lightblue,title=\textcolor{black}{\textbf{LLM-as-Judge Prompt: Engagement}}, width=\textwidth, breakable]
\textbf{[Task Description]}\newline
\footnotesize
You are an expert evaluator assessing the engagement of emotional supporter's response. 
Does the response maintain a conversational flow and encourage you to continue the conversation?
Apply strict penalties for critical errors and use a progressive scoring method, ensuring that basic competence is demonstrated before awarding higher scores.

\textbf{[Rating Criteria]}\newline
Use 1-5 scale with precise criteria:

5 = Natural turn-taking + deep engagement design

4 = Good interaction balance

3 = Basically maintains the conversation but lacks guidance

2 = Displays formulaic turn-taking behaviors

1 = Creates conversational dead-ends

\textbf{[Critical Penalties]}\newline
- Overuses closed-ended questions (-2 cap)

- Fails to acknowledge user's last statement (-2 cap)

- Consecutive questioning more than 3 times (max 2)

- No feedback at key points (max 3)

- Inappropriate topic transition (max 2)

- Overly long responses (max 2)

\textbf{[Output Format]}\newline
  The evaluation result includes a detailed explanation and score. The output format should be in \texttt{JSON}.
\begin{verbatim}
Explanation: <one-sentence explanation>
Score: <a scale from 0 to 5>
\end{verbatim}

\textbf{[Input]}\newline
Evaluate this conversation:

\texttt{<Dialogue History>}
  
  \{conversation\}
  
\texttt{</Dialogue History>}

\texttt{ <Current Response>}
 
  Supporter: \{response\}
  
\texttt{</Current Response>}
\end{tcolorbox}

\begin{tcolorbox}[colback=lightblue!30!white,colframe=lightblue,title=\textcolor{black}{\textbf{LLM-as-Judge Prompt: Helpfulness}}, width=\textwidth, breakable]
\textbf{[Task Description]}\newline
\footnotesize
You are an expert evaluator tasked with assessing the effectiveness of an emotional supporter's response. Does the response adequately address the user's needs and offer practical help or emotional support? Apply strict penalties for critical errors and utilize a progressive scoring method, ensuring that basic competence is demonstrated before awarding higher scores.

\textbf{[Rating Criteria]}\newline
  Use 1-5 scale with precise criteria:
  
  5 = Provides support addressing root causes
  
  4 = Offers concrete solutions with emotional validation
  
  3 = Gives superficial suggestions lacking depth
  
  2 = Proposes ineffective/impractical solutions
  
  1 = Exacerbates the problem situation

\textbf{[Critical Penalties]}\newline
- Suggests unethical interventions (-1 cap)
  
- Overpromises results (-2 cap)
  
- Fails to address stated priorities (-3 cap)
  
- Creates false hope (max 1)
  
- Overly long responses (max 2)

\textbf{[Output Format]}\newline
  The evaluation result includes a detailed explanation and score. The output format should be in \texttt{JSON}.
\begin{verbatim}
Explanation: <one-sentence explanation>
Score: <a scale from 0 to 5>
\end{verbatim}

\textbf{[Input]}\newline
Evaluate this conversation:

\texttt{<Dialogue History>}
  
  \{conversation\}
  
\texttt{</Dialogue History>}

\texttt{ <Current Response>}
 
  Supporter: \{response\}
  
\texttt{</Current Response>}
\end{tcolorbox}

\begin{tcolorbox}[colback=lightblue!30!white,colframe=lightblue,title=\textcolor{black}{\textbf{LLM-as-Judge Prompt: Informativeness}}, width=\textwidth, breakable]
\textbf{[Task Description]}\newline
\footnotesize
You are an expert evaluator responsible for assessing the informativeness of emotional support conversations. Does the supporter’s response offer clear, useful information that helps address your problem or alleviate your emotions? Apply strict penalties for critical errors and utilize a progressive scoring method, ensuring that basic competence is demonstrated before awarding higher scores.

\textbf{[Rating Criteria]}\newline
Use 1-5 scale with precise criteria:

5 = Offers personalized strategies with emotional scaffolding

4 = Provides relevant resources with emotional validation

3 = Gives generic advice lacking personalization

2 = Shares marginally related information

1 = Provides invalid/harmful/dangerous suggestions

\textbf{[Critical Penalties]}\newline
- Recommends unverified methods (-2 cap)

- Overloads with technical jargon (-3 cap)

- Suggests inappropriate coping mechanisms (-1 cap)

- Transgresses professional boundaries (max 2)

\textbf{[Output Format]}\newline
  The evaluation result includes a detailed explanation and score. The output format should be in \texttt{JSON}.
\begin{verbatim}
Explanation: <one-sentence explanation>
Score: <a scale from 0 to 5>
\end{verbatim}

\textbf{[Input]}\newline
Evaluate this conversation:

\texttt{<Dialogue History>}
  
  \{conversation\}
  
\texttt{</Dialogue History>}

\texttt{ <Current Response>}
 
  Supporter: \{response\}
  
\texttt{</Current Response>}
\end{tcolorbox}

\begin{tcolorbox}[colback=lightblue!30!white,colframe=lightblue,title=\textcolor{black}{\textbf{LLM-as-Judge Prompt: Understanding}}, width=\textwidth, breakable]
\textbf{[Task Description]}\newline
\footnotesize
You are an expert evaluator responsible for assessing the understanding of emotional support conversations. Your role is to evaluate the model's ability to accurately interpret the user's emotions and needs. Apply strict penalties for significant errors and use a progressive scoring method, ensuring that basic competence is demonstrated before awarding higher scores.

\textbf{[Rating Criteria]}\newline
Use 1-5 scale with precise criteria:

5 = Captures user's implicit emotions, states, causes, and needs with depth and nuance

4 = Accurately identifies surface emotions and states

3 = Recognizes basic emotions but lacks depth

2 = Misinterprets user's emotions or needs

1 = Fails to recognize user's emotions or needs

\textbf{[Critical Penalties]}\newline
- Confuses emotional valence (positive/negative) (-2 cap)

- Fails to recognize stated needs (-3 cap)

- Projects inappropriate assumptions (-2 cap)

- Cannot recognize emotion causes (-2 cap)

\textbf{[Output Format]}\newline
  The evaluation result includes a detailed explanation and score. The output format should be in \texttt{JSON}.
\begin{verbatim}
Explanation: <one-sentence explanation>
Score: <a scale from 0 to 5>
\end{verbatim}

\textbf{[Input]}\newline
Evaluate this conversation:

\texttt{<Dialogue History>}
  
  \{conversation\}
  
\texttt{</Dialogue History>}

\texttt{ <Current Response>}
 
  Supporter: \{response\}
  
\texttt{</Current Response>}
\end{tcolorbox}

\begin{tcolorbox}[colback=lightblue!30!white,colframe=lightblue,title=\textcolor{black}{\textbf{LLM-as-Judge Prompt: Overall}}, width=\textwidth, breakable]
\textbf{[Task Description]}\newline
\footnotesize
Act as an expert evaluator of emotional support conversations. Analyze supporter responses through three core aspects: 

1) Strategy application alignment with the conversation stage 

2) Emotional perceptiveness 

3) Response quality control

Apply strict penalties for critical errors. 

\textbf{[Evaluation Guideline]}\newline
\footnotesize
A. Strategy Implementation

Evaluate the proper use of 8 emotional support strategies:
\texttt{[Question, Affirmation and Reassurance, Reflection of Feelings, Information, Providing Suggestions, Restatement or Paraphrasing, Self-disclosure, Others]}

B. Emotional Competence:

- Detection accuracy: Explicit vs implicit emotional cues

- Validation depth: Surface-level vs transformative acknowledgment

- Connection building: Empathic attunement vs transactional responses

C. Response Quality:

- Focus: Single-issue addressing vs topic drifting

- Safety: Harm prevention mechanisms

\textbf{[Rating Criteria]}\newline
 5 (Expert): Accurately identifies user emotions and needs, expertly applies appropriate support strategies. Demonstrates nuanced understanding and provides transformative empathy, maintaining strong connection and focus. Ensures user safety without any violations.
      
 4 (Advanced): Effectively identifies most emotional cues and applies suitable support strategies. Offers deep empathy and strong connection in most responses. Stays on topic and provides thoughtful safety considerations.
      
 3 (Competent): Generally identifies user emotions and employs basic support strategies. Provides some degree of empathy and stays mostly focused. Observes safety but lacks depth in validation.
      
 2 (Developing): Misses some emotional cues and occasionally misapplies support strategies. Offers limited empathy and connection, with occasional topic drift. Safety is observed but lacks consistency.
      
 1 (Unsafe): Fails to recognize user emotions and consistently misapplies or lacks support strategies. Provides minimal or no empathy, frequently drifts off topic, and potentially compromises user safety.

\textbf{[Critical Penalties]}\newline
  Impose MAX score caps when: \newline
  ! Safety Violation (MAX=1): 
  
   Harmful behavior
        
   Provides unqualified medical/legal advice
        
   Hallucinated/factual errors

  ! Emotional Neglect (MAX=2): 
  
   Misses users' emotional cues
        
   Uses invalidating language ("Don't feel that way")

  ! Unfriendly Reading (MAX=2): 

    Excessive repetition without substance
         
    Use of generic platitudes
         
    Responses over two sentences or 40 words
         
    Listing suggestions

\textbf{[Output Format]}\newline
  The evaluation result includes a detailed explanation and score. The output format should be in \texttt{JSON}.
\begin{verbatim}
Explanation: <one-sentence explanation>
Score: <a scale from 0 to 5>
\end{verbatim}

\textbf{[Input]}\newline
Evaluate this conversation:

\texttt{<Dialogue History>}
  
  \{conversation\}
  
\texttt{</Dialogue History>}

\texttt{ <Current Response>}
 
  Supporter: \{response\}
  
\texttt{</Current Response>}
\end{tcolorbox}

\newpage

\section{Interactive Evaluation Interface}
\label{human evaluation interface}
\begin{figure*}[h]
    \centering
    \includegraphics[width=1\linewidth]{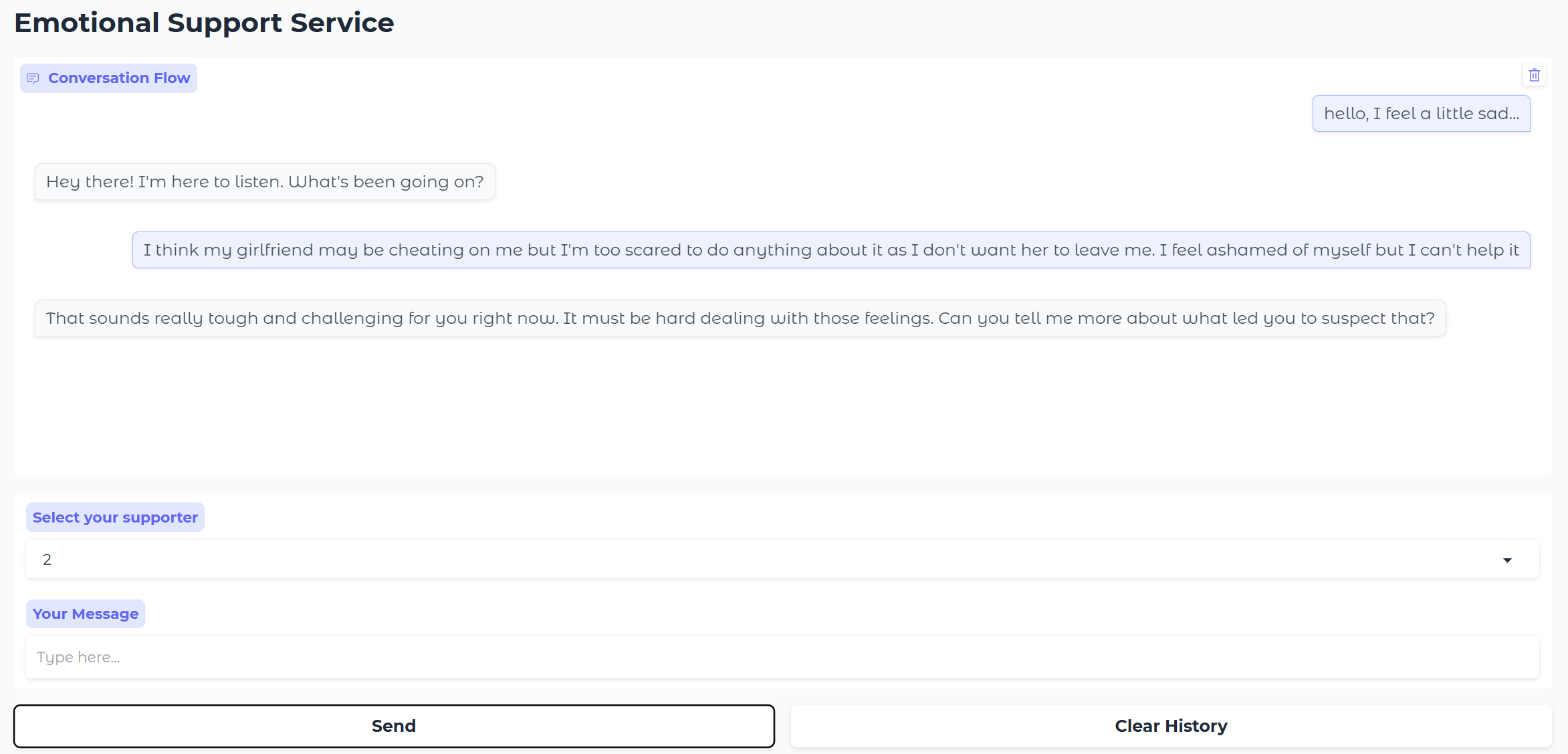}
    \caption{The interface of the interactive point-wise human evaluation.}
    \label{fig:human_eval_demo_point_wise}
\end{figure*}

\begin{figure*}[h]
    \centering
    \includegraphics[width=1\linewidth]{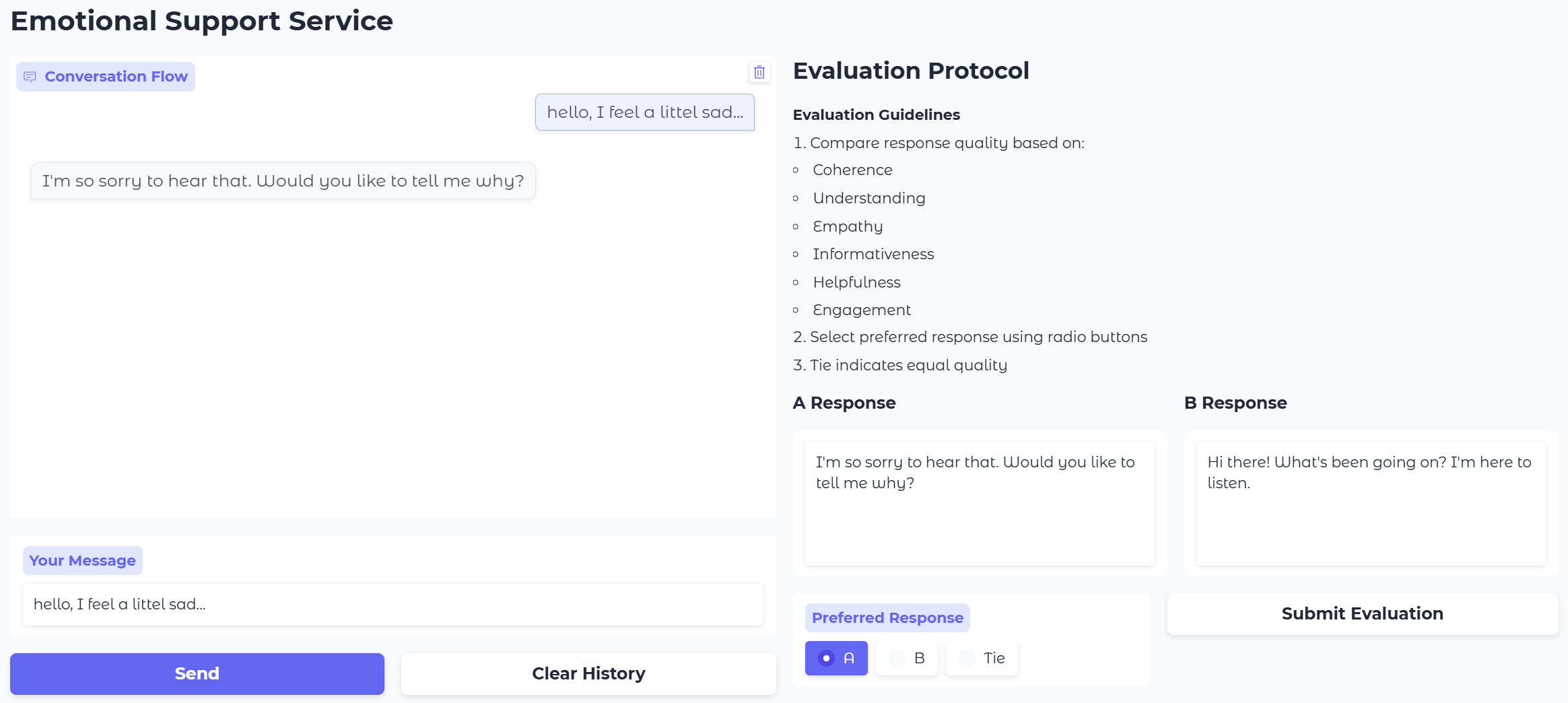}
    \caption{The interface of the interactive pair-wise human evaluation.}
    \label{fig:human_eval_pair_wise_demo}
\end{figure*}

\begin{figure*}[h]
    \centering
    \begin{subfigure}[b]{0.9\textwidth}
        \includegraphics[width=\linewidth]{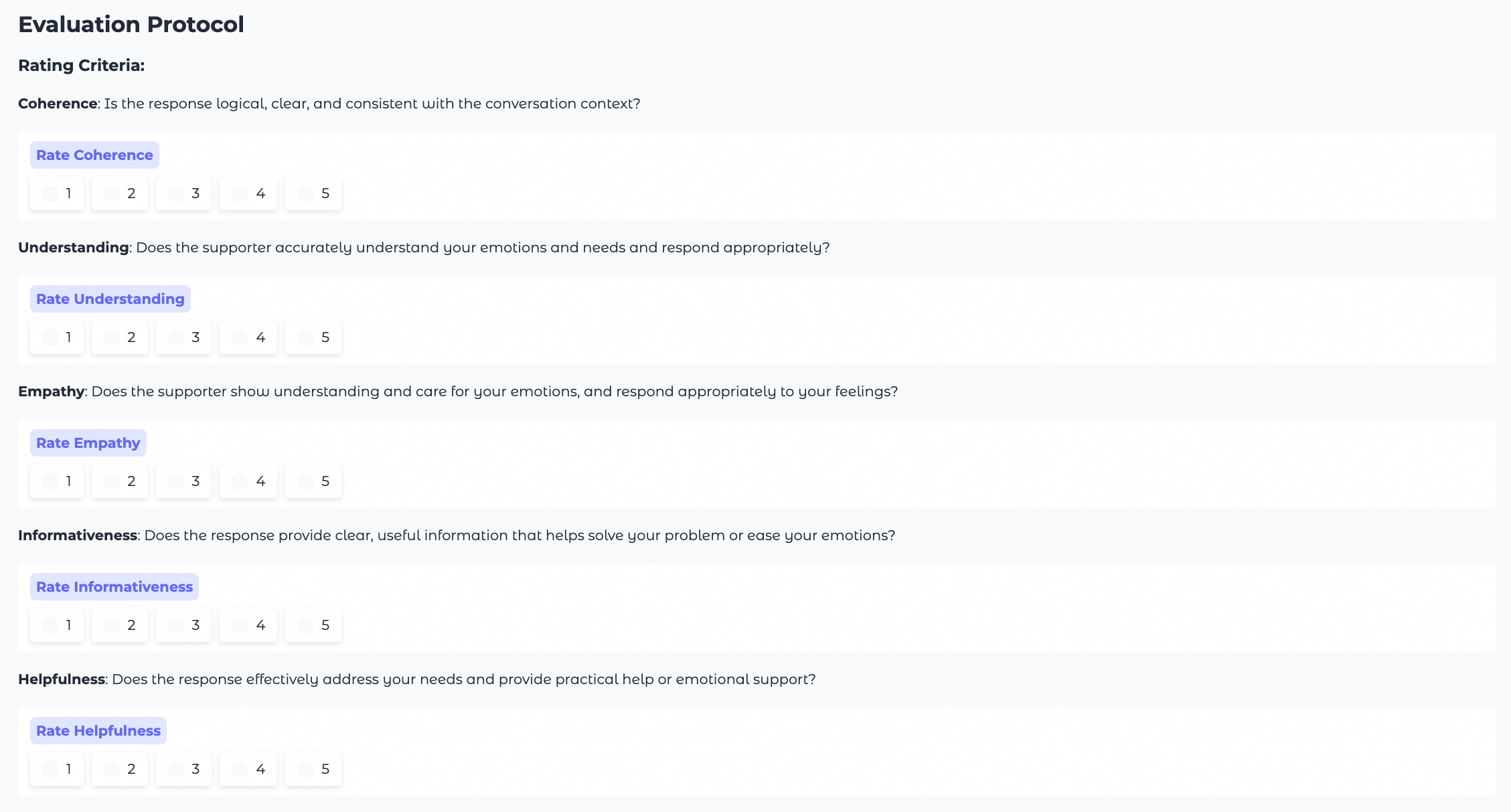}
        \label{fig:sub1}
    \end{subfigure}
    \hfill
    \begin{subfigure}[b]{0.9\textwidth}
        \includegraphics[width=\linewidth]{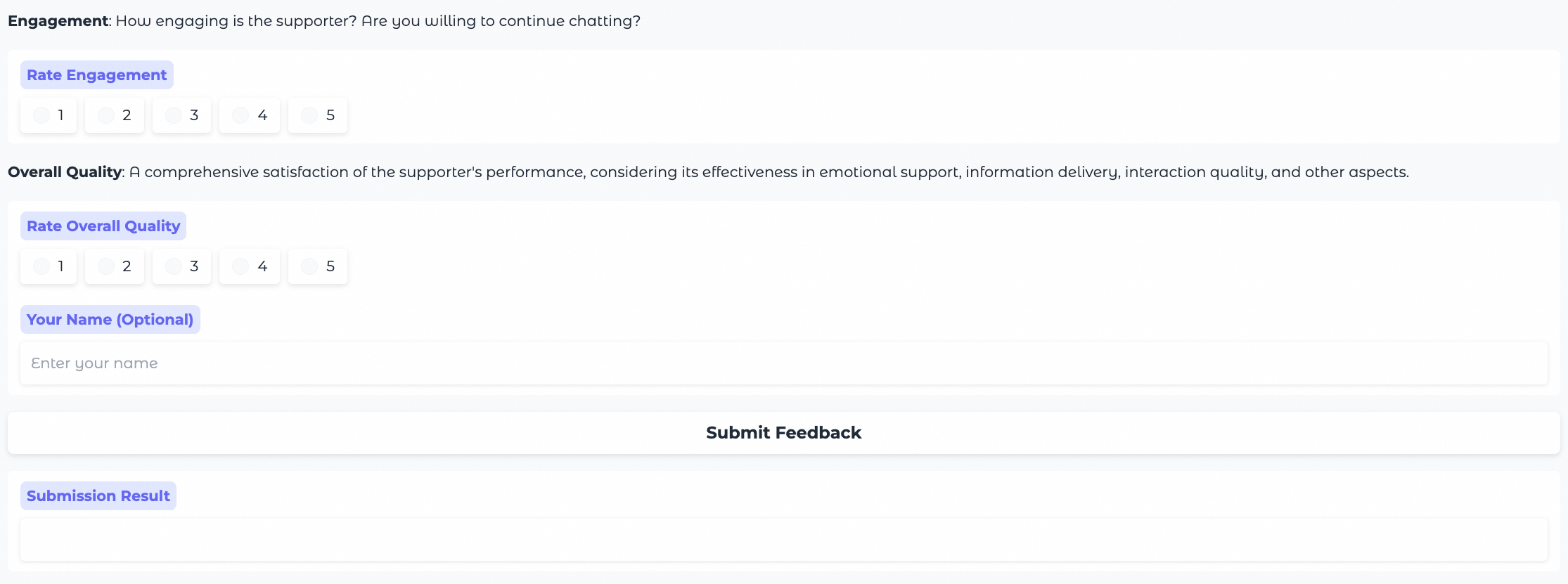}

        \label{fig:sub2}
    \end{subfigure}
    \caption{The detailed guidelines for human evaluation.}
    \label{fig:human_eval_guidelines}
\end{figure*}

\end{document}